\documentclass[sigconf]{acmart}

\usepackage{subfigure}
\usepackage{nicematrix}
\usepackage[ruled, linesnumbered]{algorithm2e}
\usepackage{bbm}
\usepackage{enumitem}
\usepackage{booktabs}

\AtBeginDocument{%
  \providecommand\BibTeX{{%
    \normalfont B\kern-0.5em{\scshape i\kern-0.25em b}\kern-0.8em\TeX}}}

\copyrightyear{2025}
\acmYear{2025}
\setcopyright{acmlicensed}
\acmConference[KDD '25] {Proceedings of the 31st ACM SIGKDD Conference on Knowledge Discovery and Data Mining}{August 3--7, 2025}{Toronto, ON, Canada.}
\acmBooktitle{Proceedings of the 31st ACM SIGKDD Conference on Knowledge Discovery and Data Mining (KDD '25), August 3--7, 2025, Toronto, ON, Canada}
\acmISBN{979-8-4007-1245-6/25/08}
\acmDOI{10.1145/3690624.3709239} 

\begin{document}

\title{Augmented Contrastive Clustering with Uncertainty-Aware Prototyping for Time Series Test Time Adaptation}

\author{Peiliang Gong}
\affiliation{%
  \institution{Nanjing University of Aeronautics and Astronautics}
  \city{Nanjing}
  \country{China}
}
\email{plgong@nuaa.edu.cn}

\author{Mohamed Ragab}
\affiliation{%
  \institution{Technology Innovation Institute}
  \city{Masdar}
  \country{United Arab Emirates}}
\email{mohamedr002@e.ntu.edu.sg}

\author{Min Wu}
\affiliation{%
  \institution{Institute for Infocomm Research, Agency for Science Technology and Research (A*STAR)}
  \country{Singapore}
}
\email{wumin@i2r.a-star.edu.sg}

\author{Zhenghua Chen}
\affiliation{%
 \institution{Institute for Infocomm Research, Agency for Science Technology and Research (A*STAR)}
 \country{Singapore} \\
 \institution{Centre for Frontier AI Research, Agency for Science Technology and Research (A*STAR)}
 \country{Singapore}
}
\email{chen0832@e.ntu.edu.sg}

\author{Yongyi Su}
\affiliation{%
  \institution{South China University of Technology}
  \city{Guangzhou}
  \country{China} \\
  \institution{Institute for Infocomm Research, Agency for Science Technology and Research (A*STAR)}
 \country{Singapore}
}
\email{su.yongyi.syy@gmail.com}

\author{Xiaoli Li}
\authornote{Corresponding Author}
\affiliation{%
 \institution{Institute for Infocomm Research, Agency for Science Technology and Research (A*STAR)}
 \country{Singapore} \\
 \institution{Centre for Frontier AI Research, Agency for Science Technology and Research (A*STAR)}
 \country{Singapore}
}
\email{xlli@i2r.a-star.edu.sg}

\author{Daoqiang Zhang}
\authornotemark[1]
\affiliation{%
  \institution{Nanjing University of Aeronautics and Astronautics}
  \city{Nanjing}
  \country{China}}
\email{dqzhang@nuaa.edu.cn}

\renewcommand{\shortauthors}{Peiliang Gong et al.}

\begin{abstract}
Test-time adaptation aims to adapt pre-trained deep neural networks using solely online unlabelled test data during inference. Although TTA has shown promise in visual applications, its potential in time series contexts remains largely unexplored. Existing TTA methods, originally designed for visual tasks, may not effectively handle the complex temporal dynamics of real-world time series data, resulting in suboptimal adaptation performance. To address this gap, we propose Augmented Contrastive Clustering with Uncertainty-aware Prototyping (ACCUP), a straightforward yet effective TTA method for time series data. Initially, our approach employs augmentation ensemble on the time series data to capture diverse temporal information and variations, incorporating uncertainty-aware prototypes to distill essential characteristics. Additionally, we introduce an entropy comparison scheme to selectively acquire more confident predictions, enhancing the reliability of pseudo labels. Furthermore, we utilize augmented contrastive clustering to enhance feature discriminability and mitigate error accumulation from noisy pseudo labels, promoting cohesive clustering within the same class while facilitating clear separation between different classes. Extensive experiments conducted on three real-world time series datasets and an additional visual dataset demonstrate the effectiveness and generalization potential of the proposed method, advancing the underexplored realm of TTA for time series data.
\end{abstract}



\maketitle

\section{Introduction}
Real-world time series applications frequently suffer from performance degradation due to domain shifts between training and test data. Unsupervised domain adaptation (UDA) methods \cite{UDA} have emerged to mitigate this impact by leveraging unlabeled target data, aiming to align feature representations in the embedding space \cite{UDA_TS1, UDA_TS2, UDA_TS3, UDA_TS4, UDA_TS5, UDA_TS6, adatime}. However, these approaches typically require access to source domain data during adaptation, which may be infeasible due to privacy and security concerns.
\begin{figure}[t]
\centering
\setlength{\subfigcapskip}{5pt} 
\begin{tabular}{cc}
\subfigure[]{\includegraphics[width=0.49\linewidth]{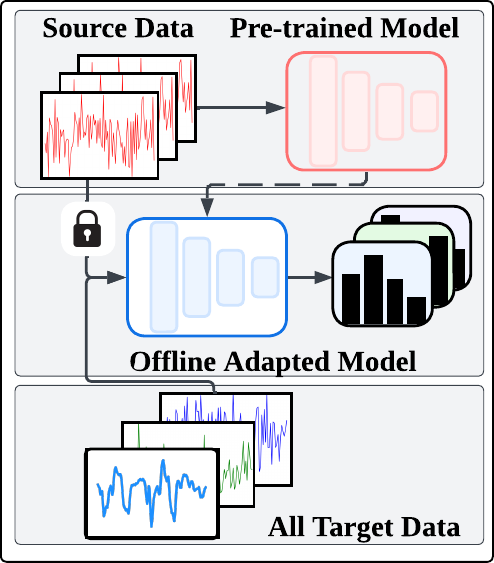} \label{figure1_a}}
\subfigure[]{\includegraphics[width=0.49\linewidth]{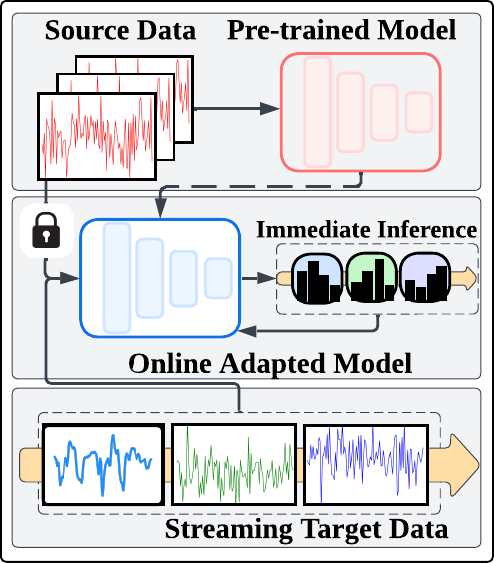} \label{figure1_b}}
\end{tabular}
\caption{Illustration of SFDA and TTA for time series applications. (a) SFDA utilizes all available target data for multi-epoch adaptation before making final predictions. (b) TTA adapts the pre-trained model to target data batches in an online manner, where each batch can only be observed once.}
\label{figure1}
\end{figure}
Source-free domain adaptation (SFDA) offers an alternative by handling domain shifts without source data, demonstrating promising results by considering temporal dynamics \cite{SFDA, mapu, SFDA_TS, SFDA_TS2}. Nonetheless, SFDA struggles in real-world scenarios, particularly online time series applications where continuous data streams demand immediate inference. For instance, deploying a pre-trained fault diagnosis model for real-time monitoring on production lines using SFDA becomes impractical due to its requirement for pre-collecting all target data, rendering it unsuitable for on-the-fly diagnostics.

Test-time adaptation (TTA) provides a more viable solution, enabling on-the-fly model adaptation with incoming data streams \cite{TTA_survey}. This approach circumvents the high training costs associated with SFDA and ensures the model remains responsive to new information, ideal for applications requiring immediate inference, as illustrated in Figure \ref{figure1}. Recent advancements in TTA for various vision applications have shown promising progress \cite{TTT, TTT++, TDA, TRIBE, MEMO, LAME, SoTTA, AaDNPC}. Nevertheless, these approaches are primarily designed for visual tasks and may overlook the critical temporal characteristic inherent in real-world time series scenarios. Consequently, directly applying existing methods to time series data often yields suboptimal performance.

Time series data present distinct challenges compared to other data types, particularly due to magnitude variability and inherent noise. Magnitude variability arises when the amplitude of time series data changes due to factors such as differing operational conditions, sensor calibrations, or individual differences. For instance, in human activity recognition, the same activity, such as walking, can produce varying magnitudes in accelerometer readings depending on the individual's walking speed, requiring the model to accurately classify the activity regardless of these amplitude variations. Additionally, real-world time series data are often noisy and non-stationarity, with external interferences or environmental factors obscuring true patterns and complicating accurate data interpretation. These challenges are particularly pronounced under the test-time adaptation settings, where the model must adapt instantaneously to the evolving temporal dynamics of the target data without access to the source domain. Therefore, our key question is \textit{how to effectively alleviate the negative impact of these temporal characteristic during the test time adaptation phase.}

To overcome these challenges, we propose Augmented Contrastive Clustering with Uncertainty-aware Prototyping (ACCUP), a novel TTA method for time series data. Our solution introduces an uncertainty-aware prototypical ensemble module that enhances the reliability of model outputs. First, this module employs magnitude warping augmentation to generate amplitude variations in the time series data, enabling the model to learn temporal patterns that are invariant to these changes and enrich feature representations with more informative content. Uncertainty-aware prototypes are then calculated using entropy-minimized ensemble predictions from the augmentation-enhanced features. By focusing on these most confident and stable temporal patterns, it mitigates the effects of noise and distribution shifts. In addition, an entropy comparison scheme further ensures trustworthy predictions by filtering out instances with high uncertainty estimates.

Despite its effectiveness, the inevitable occurrence of some noisy pseudo-labels can diminish model performance through error propagation. Using backpropagation with cross-entropy loss on noisy labels can mislead the training process, detrimentally affecting the model's adaptation \cite{cross_entropy_reason1, cross_entropy_reason2, cross_entropy_reason3}. To mitigate this, we further introduce an augmented contrastive clustering strategy to enhance feature discriminability and reduce the influence of noisy pseudo-labels by prioritizing reliable predictions. By maximizing mutual information between raw and augmented data, the model maintains temporal coherence and ensures consistent learned temporal dependencies across different views. In addition, augmented contrastive clustering also promotes tight grouping of instances within the same class and clear separation between different classes, ensuring that the temporal patterns unique to each class are preserved and accurately classified. This is particularly crucial for time series data, where subtle temporal variations often define class boundaries.

The main contributions of this paper can be summarized below:
\begin{itemize}
    \item To our knowledge, this is the first work to advance TTA for a broad spectrum of real-world time series applications, despite the existence of some domain-specific approaches.
    \item We propose a novel uncertainty-aware prototypical ensemble module designed to prioritize confident temporal patterns and rectify the quality of pseudo-labels, thereby enhancing their trustworthiness.
    \item We present an augmented contrastive clustering constraint to improve feature discriminability and mitigate error accumulation during model adaptation.
    \item Extensive comparative and ablation experiments demonstrate the effectiveness of our proposed approach in multiple real-world time series applications.
\end{itemize}

\begin{figure*}[t]
\centering
\includegraphics[width=0.95 \linewidth]{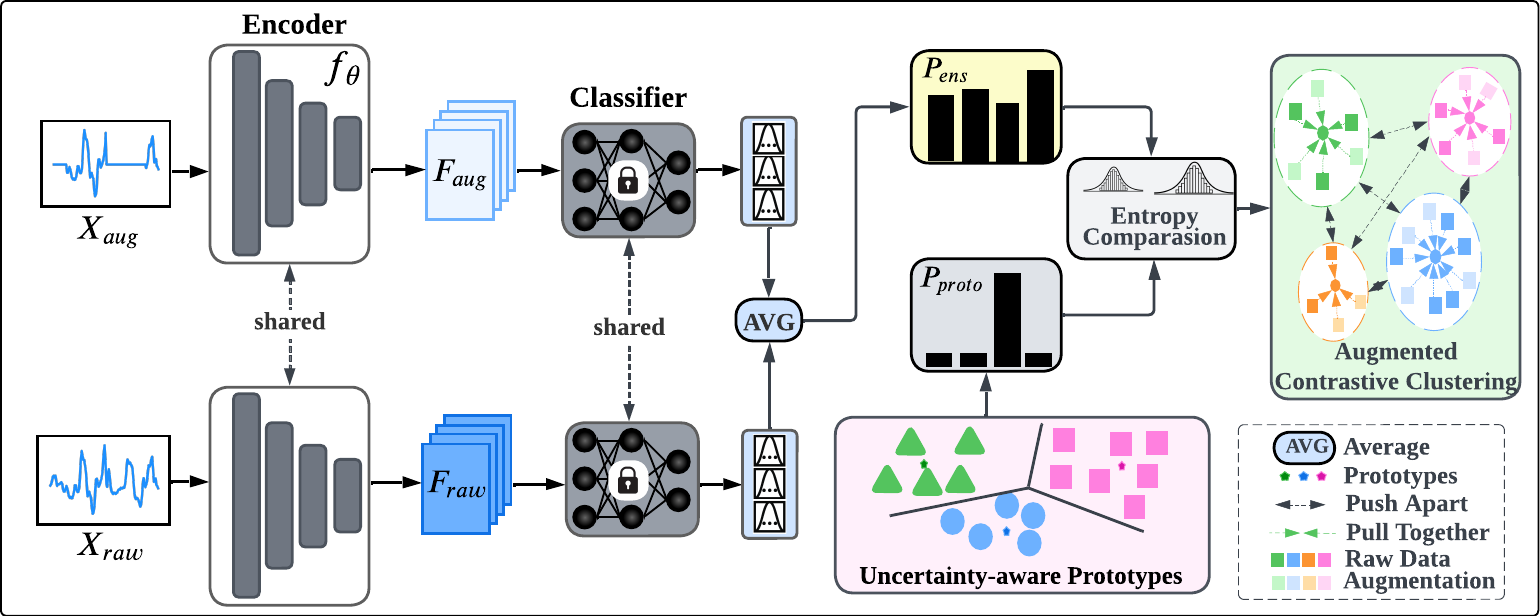}
\caption{Illustration of the proposed model. Given the input signal, initial enhanced features and predictions are obtained via augmentation-ensemble, and then the predictions are further corrected by uncertainty-aware prototypes. Further, trustworthy pseudo-labels are obtained using an entropy comparison scheme. Lastly, more discriminative features are learnt using augmented contrastive clustering constraints to mitigate error accumulation and improve model efficacy.}
\label{figure2}
\end{figure*}

\section{Related Works}
\subsection{Time Series Domain Adaptation}
UDA was developed to bridge the performance gap in deploying models across different domains without target annotations. Significant strides in UDA for time series tasks have emerged through two main strategies: discrepancy aligning-based methods and adversarial learning \cite{adatime}. Discrepancy aligning aims to minimize statistical differences for domain alignment \cite{UDA_TS6, UDA_TS1, UDA_Align}, while adversarial learning promotes domain-invariant features through adversarial training \cite{UDA_TS2, UDA_TS3, UDA_TS4, UDA_TS5}. For instance, RAINCOAT tackles feature and label shifts by aligning temporal and frequency features across domains \cite{UDA_TS6}, while SLARDA utilizes autoregressive adversarial training to align temporal dynamics \cite{UDA_TS3}. Despite these methods showing promising results, they rely on access to source data during adaptation. In practical scenarios, however, obtaining source data can be challenging due to privacy concerns or storage limitations. SFDA was proposed to address this challenge \cite{mapu, SFDA_TS, SFDA_TS2}. Methods like MAPU utilize temporal imputation tasks to recover the original signal in feature space without source data \cite{mapu}. However, SFDA requires pre-existing access to all target data, and the model needs iterative training on this data, hindering efficiency. Real-world time series applications demand immediate inference and adaptation upon data arrival, creating a need for more realistic adaptation scenarios. Some recent work has made preliminary attempts to adapt the model during testing of tasks such as automatic speech recognition or video-based action recognition \cite{sgem, tta_speech, video_tta, NOTE, TTA_AD}. Although these existing methods also consider data of a sequence nature, they are considered to be application-specific and may not generalize well across applications. In addition, video and time series, while sharing some commonalities, are usually treated as different modalities. Therefore, this motivates us to explore versatile TTA methods suitable for time series data, which is underexplored.

\subsection{Test-Time Adaptation}
Test-time adaptation (TTA) presents a realistic and challenging scenario by adapting pre-trained models to online target data streams during testing. One paradigm involves incorporating auxiliary self-supervised tasks during both training and testing, which may not be feasible or scalable in real-life scenarios \cite{TTT, TTT++}. Another prevailing paradigm adapts the model solely during the testing phase using pre-trained source models and online unlabeled data \cite{BN, TENT, GDA, Rdumb, PETAL, conjugateTTA, AdaContrast}. For instance, normalization-based approaches either replace the training model's statistics with estimates from test data or adjust its normalization layer parameters \cite{BN, TRIBE, RoTTA}. Entropy-minimizing self-training enhances model generalization by minimizing entropy to adjust the trainable parameters \cite{SHOT_IM, TENT, EATA, SAR, CoTTA}. Additionally, some methods enhance adaptation performance using strategies such as feature alignment or prototype learning \cite{TTAC, T3A, TAST, TSD}. Although existing TTA methods show promise, their design primarily targets visual tasks, which may limit their effectiveness in handling the inherent noise and magnitude variability of real-world time series data. In contrast, our approach specifically addresses these limitations through an uncertainty-aware prototypical ensemble and an augmented contrastive clustering strategy, effectively boosting adaptation performance in this unique domain. 

\subsection{Prototypical Learning}
Prototype learning represents a paradigm where classes or concepts are characterized by central or typical examples within the feature space. It has evolved from its early combination with nearest neighbor rules for classification tasks \cite{prototype_early} to its recent prominence with the advent of deep learning. These methods have been successfully applied in various applications such as few-shot learning \cite{prototype_FS1, prototype_FS2, prototype_FS3, prototype_FS4} and continual learning \cite{prototype_CL1, prototype_CL2, prototype_CL3, prototype_CL4, prototype_CL5, prototype_CL6}. For instance, prototypical networks calculate class prototypes as mean embeddings of few-shot examples, simplifying classification by comparing new instances to these prototypes \cite{prototype_FS1}. In continual learning, CoPE \cite{prototype_CL2} extends this concept by incorporating a high momentum-based update strategy for prototypes with each observed batch. Additionally, substantial research has focused on using prototype learning to create compact feature spaces for open-set recognition \cite{prototype_OS1, prototype_OS2, prototype_OS3, prototype_OS4, prototype_OS5} and to enhance model interpretability \cite{prototype_Int1, prototype_Int2}. Recent advancements have explored dynamic prototype updates during testing \cite{T3A, TSD, TAST}, as exemplified by T3A \cite{T3A}, which updates prototypes using a support set comprising original classifier weights and online data features based on pseudo-labels. However, these methods often struggle to maintain prototype reliability in the presence of noisy data, particularly in time-series domains where temporal variations can introduce significant challenges.

\section{Methodology}
\subsection{Problem Definition}
Given the pre-trained model with standard empirical risk minimization on the source $\mathcal{D}_S = \{ X_S^i, y_S^i \}_{i=1}^{n_S}$, our goal is to adapt the source pre-trained model using unlabelled target data $ \mathcal{D}_T = \{ X_T^i \}_{i=1}^{n_T}$. Here, $X_S$ and $X_T$ denote univariate or multivariate time series of length $L$ in the source and target domains, respectively, $y_S$ denotes the category labels corresponding to $X_S$, $n_S$ and $n_T$ denote the number of source and target samples, respectively. During testing, we initialize model $g= h_{\phi} \circ f_{\theta}$ using the source pre-trained model parameters, where $f_{\theta}$ denotes the feature encoder and $h_{\phi}$ denotes the linear classifier. The output of model $g$ for time series $X_T^i$ is written as $p_i = g(X_T^i) \in \mathbb{R}^C$, where $C$ is the number of categories. This work aims to address the test-time adaptation problem, where the source-domain data is strictly inaccessible and the training process of the source-domain model is fixed. This assumption is more realistic in real-world application scenarios, taking into account data privacy concerns as well as the direct deployment of source-domain pre-trained models. In addition, we assume a difference across the marginal distributions, i.e., $P(X_S) \neq P(X_T)$, while the conditional distributions are stable, i.e., $P(y_S|X_S) \approx P(y_T|X_T)$.

\subsection{Overview}
Our proposed method, ACCUP, is designed for real-world time series adaptation. Figure \ref{figure2} illustrates its main pipeline. The model includes two key modules: (1) uncertainty-aware prototypical ensemble first enhances predictions through magnitude warping augmentation and then refines them using uncertainty-aware prototypes. Finally, it generates reliable pseudo-labels based on an entropy comparison scheme, effectively addressing the magnitude variability and noise inherent in time series data; (2) the augmented contrastive clustering mitigates error accumulation of noisy pseudo-labels by simultaneously considering cross-view prediction consistency, sample cohesion within classes, and clear separation between classes. It ensures robustness against test-time variations, maintaining class conditional invariance despite the challenges posed by time series data. We will elaborate on each component in the next subsections. The complete algorithm is provided in the Appendix \ref{algorithm_1}.

\subsection{Uncertainty-Aware Prototypical Ensemble}
\subsubsection{Augmentation-Ensemble Predictions}
Time series often exhibit significant variability in amplitude due to various factors such as operational conditions or environmental influences. This variability can cause substantial discrepancies between training and test data, making it difficult for models to generalize and perform accurately on new data. To mitigate the challenges posed by this characteristic, we introduce an augmentation-ensemble strategy for refined temporal features and logits in the context of TTA. However, unlike standard augmentation during training, TTA involves fine-tuning unlabeled target data during inference. Aggressive augmentation might deviate significantly from the target distribution, leading to inconsistent predictions and hindering performance. It is essential to choose a moderate augmentation strategy so that the generated views can retain overall the intrinsic characteristics of the original time series while still introducing enough variability. To cater for this, we introduce magnitude warping augmentation to alleviate magnitude variability by integrating information from both raw and augmentated views \cite{magnitude_warp_aug}. Magnitude warping involves stretching or compressing the magnitude of the time series signal by convolving the data window with a smooth curve varying around one. This augmentation method performs elastic transformations that introduce nuanced variations while ensuring that the innate temporal dynamics of the time series data are preserved.

Given a batch of unlabelled test samples $\mathcal{X}$, we use the magnitude warping to obtain the corresponding augmented views $\mathcal{A}$. For each time series $x \in \mathcal{X}$ and $a \in \mathcal{A}$, we can obtain the corresponding feature embeddings and predicted logits: $f_{raw} = f_{\theta}(x)$, $p_{raw} = h_{\phi}(f_{raw})$, $f_{aug} = f_{\theta}(a)$, and $p_{aug} = h_{\phi}(f_{aug})$, respectively. Thus augmentation-ensemble features and logits are then derived by averaging, generating more robust representations as follows,
\begin{equation}
    \begin{split}
    f_{ens} &= (f_{raw} + f_{aug}) / 2, \\
    p_{ens} &= (p_{raw} + p_{aug}) / 2.
    \end{split}
\end{equation}

For simplicity, a simple averaging ensemble of the two views is used here. To further investigate the impact of different weighting schemes, additional experiments are conducted as presented in the Appendix \ref{appendix_different_weight_ensemble}.

\begin{figure}[t]
\centering
\includegraphics[width = 0.45 \textwidth]{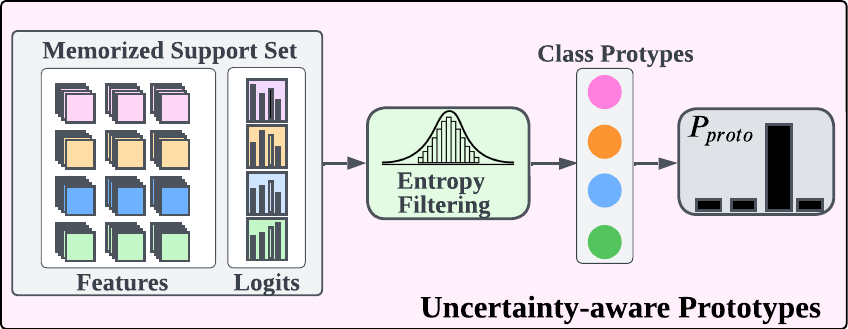} 
\caption{Illustration of the uncertainty-aware prototypes. The memorized support set is first initialized using the weight parameters of the linear classifier from the pre-trained source model. When the target domain batch of data arrives, the enhanced features and logits obtained from the augmentation-ensemble are deposited into the memorized support set, and then entropy filtering is used to select plausible representatives of the support set to produce a more reliable class prototypes.}
\label{figure3}
\end{figure}

\subsubsection{Uncertainty-Aware Prototypes}
Time series data are typically noisy and subject to distribution shifts between training and testing phases. To prioritize more confident temporal patterns and mitigate the impact of noise, we present uncertainty-aware prototypes in this context by incorporating uncertainty estimates to weigh features based on their reliability, as shown in Figure \ref{figure3}. Prototypes often are less sensitive to outliers and treat categories equally, addressing class-imbalanced real-world time-series data. Specifically, since labeled data is not available in TTA, we compute pseudo-class prototypes using the memorized support set that stores the past test data and the corresponding logits. The memorized support set starts with the pre-trained source model's linear classifier weights. When a batch of target-domain data $\mathcal{X}$ arrives, we add the feature embeddings $f_{ens}$ and logits $p_{ens}$ of each sample $x \in \mathcal{X}$ obtained from the augmentation-ensemble module to the memorized support set. To ensure reliable prototypes, we filter out unreliable using Shannon entropy \cite{Shannon_entropy}. Specifically, the entropy of the predicted logits $p_{ens}$ is $H_{ens} = -\sum_{c=1}^C \sigma(p_{ens}^c) \cdot {\rm log} \, \sigma(p_{ens}^c)$, where $\sigma$ denotes the softmax activation function. For each category, the $K$ samples with the lowest entropy (most confident predictions) are retained. Each category's prototype is then computed as the centroid of the filtered feature embeddings. Finally, the uncertainty-aware prototype $\mu_c$ for class $c$ can be calculated as:
\begin{equation}
    \mu_{c} = \frac{\sum_{i} f_{ens}^i \mathbb{I}[\hat{y_i}=c]}{\sum_{i} \mathbb{I}[\hat{y_i}=c]},
\end{equation}
where $\hat{y_i} = {\rm arg \, max} \, p_{ens}^i$ is pseudo label of the current sample and $\mathbb{I}[\cdot]$ denotes the indicator function which outputs $1$ when argument satisfies the condition otherwise outputs $0$.  We determine each sample's class prediction by calculating the distance between its feature embedding and the prototype of its corresponding category. Formally, for the current batch of samples $x \in \mathbb{B}$, we obtain the prototype-based classification logits for category $c$ as follows:
\begin{equation}
    p_{proto}^c = \frac{{\rm exp}(sim(f_{ens}, \mu_{c}) * \eta)}{\sum_{c'=1}^C {\rm exp}(sim(f_{ens}, \mu_{c'}) * \eta)},
\end{equation}
where $\eta$ is the scaling parameter, $sim(f_{ens}, \mu_c)$ denotes cosine similarity between two feature embeddings. 

\subsubsection{Entropy Comparison Scheme}
To rectify the quality of the pseudo-labels used for the augmented contrastive clustering, we further propose an entropy comparison scheme. This scheme selectively combines predictions from the augmented-ensemble module and the uncertainty-aware prototype classification based on their respective entropy values. For each sample $x$, the prediction with more credibility (i.e., lower entropy) is chosen as the final output logits, which denoted as,
\begin{equation}
    p_{out} = p_{ens} \cdot \mathbb{I}\left[H_{ens}<H_{proto}\right]+ p_{proto} \cdot \mathbb{I}
    \left[H_{ens}\geq H_{proto}\right].
\end{equation}

Here, $\mathbb{I}\left[\cdot \right]$ is the indicator function, and $H_{ens}$ and $H_{proto}$ are the Shannon entropies of $p_{ens}$ and $p_{proto}$, respectively. Subsequently, the pseudo-label for $x$ is obtained as $\hat{y} = {\rm arg \, max} \, p_{out}$.

\subsection{Augmented Contrastive Clustering}
Although securing reliable pseudo-labels significantly boosts performance, self-training with cross-entropy remains sensitive to noisy labels, where even a small number can detrimentally harm performance \cite{cross_entropy_reason1, cross_entropy_reason2, cross_entropy_reason3}. To address this issue and enhance feature discrimination, we propose an augmented contrastive clustering to optimize our adaptation strategy. Instead of relying on cross-entropy which directly propagates labels, contrastive cluster is based on similarity rather than explicit label assignment, which inherently mitigates the risk of propagating erroneous labels. Even if some labels within a cluster are noisy, their influence is diluted as the model is primarily guided by the majority of correctly labeled examples in each cluster. Specifically, our module's design focuses on two key aspects. First, ensuring that predictions from both raw and augmented data have similar distributions is vital for maintaining the temporal coherence of time series data. By maximizing mutual information between similar temporal patterns across different views, the model can better capture the underlying temporal dynamics. Second, temporal data often require distinguishing between subtle temporal patterns, which can be obscured by noise. By promoting tight clustering within the same class and clear separation between different classes, this method ensures that the temporal dependencies are preserved and accurately classified.

To accomplish this, we initially combine the raw time series $\mathcal{X}$ with its augmented views $\mathcal{A}$ from the current test batch to form a unified set $\mathcal{C} = \mathcal{X} \cup \mathcal{A}$. With this combined set $\mathcal{C}$, we identify positive and negative pairs for an anchor sample $x_i$ as ${\rm pos}(i) = { x_j \in \mathcal{C}; y_j = y_i }$ and ${\rm neg}(i) = { x_k \in \mathcal{C}; y_k \ne y_i }$, respectively. Given the absence of true target labels, we use pseudo-labels generated by the uncertainty-aware prototype ensemble module to categorize the samples into positive and negative sets for clustering purposes. Our objective is to maximize mutual information among positive pairs while reducing their similarity with negative pairs. To ensure consistent labeling across views, both original and augmented versions of a sample are assigned the same pseudo-label.
Finally, the augmented contrastive clustering constraint for sample $x_i$ can be defined as follows:
\begin{equation}
    \mathcal{L}_{contrast}^i = \frac{-1}{\left \vert {\rm pos}(i) \right \vert} \sum_{j \in {\rm pos}(i)}\left({\rm log} \frac{{\rm exp}({\sigma (p_i, p_j) / \tau})}{\sum_{k \in {\rm neg}(i)} {\rm exp}(\sigma(p_i, p_k) / \tau)}\right),
\end{equation}
where $\left \vert {\rm pos}(i) \right \vert$ denotes the cardinality of the positive sample set, $\tau$ is the temperature parameter to adjust the contrastive scale, and $\sigma(p_i, p_j)$ denotes cosine similarity score between the uncertainty-aware prototypical ensemble logits of two samples.

\section{Experiment and Results}
In this section, we systematically evaluate our method against state-of-the-art approaches across various time-series applications through rigorous experimentation. Further, we perform multiple ablation experiments to verify the effectiveness of the different components. Ultimately, we conduct a series of model analyses including parameter sensitivity analysis and visualization of t-SNE feature distributions. The experimental implementation details are presented in Appendix \ref{implementation_details}.

\subsection{Datasets}
\begin{table*}[h]
\centering
\caption{Detailed results of the five UCIHAR cross-domain scenarios in terms of macro F1 score.}
\setlength{\tabcolsep}{2.5mm}{
\begin{NiceTabular}{@{}l|ccccc|c@{}} 
\toprule 
Algorithm & 2$\rightarrow$11 & 6$\rightarrow$23 & 7$\rightarrow$13 & 9$\rightarrow$18 & 12$\rightarrow$16 & Average\\ \midrule
Target  & 100.0$\pm$0.0 & 100.0$\pm$0.0 & 100.0$\pm$0.0 & 100.0$\pm$0.0 &  99.50$\pm$0.43 & 99.90 \\
\midrule
Source & 99.11$\pm$1.26 & 92.28$\pm$1.85 & 95.59$\pm$0.38 & 75.38$\pm$2.57 & 58.69$\pm$1.31 &   84.21 \\
BN \cite{BN} & 98.67$\pm$0.98	& 91.91$\pm$1.95 & 96.24$\pm$0.30 & 81.97$\pm$3.33 & 60.06$\pm$1.12	& 85.77 \\
TENT \cite{TENT} & 98.67$\pm$0.98 & 91.99$\pm$1.91 & 96.24$\pm$0.30 & 81.95$\pm$3.30 & 60.15$\pm$1.14 & 85.80 \\
PL \cite{PL} & 98.67$\pm$0.98 & 92.91$\pm$1.55 & 96.67$\pm$0.15 & 81.83$\pm$4.76 & 59.78$\pm$0.93 & 85.97  \\
SHOT-IM \cite{SHOT_IM} & 98.77$\pm$1.73 & \underline{95.21$\pm$0.24} & 96.33$\pm$0.14 & 79.09$\pm$2.19 & 61.15$\pm$3.15 & 86.11  \\
EATA \cite{EATA} & 98.78$\pm$1.03 & 91.34$\pm$2.33 & 96.68$\pm$0.40 & 82.71$\pm$3.78 & 60.36$\pm$1.30 & 85.97 \\
NOTE \cite{NOTE} & \underline{99.33$\pm$0.27} & 89.60$\pm$3.85 & 96.76$\pm$0.27 & \underline{85.60$\pm$3.43} & \underline{61.17$\pm$2.88} & 86.69 \\
CoTTA \cite{CoTTA} & 98.89$\pm$1.10 & 91.00$\pm$2.39 & 96.34$\pm$0.15 & 81.67$\pm$3.77 & 59.85$\pm$1.36 & 85.55 \\
SAR \cite{SAR} & 98.78$\pm$1.03 & 91.43$\pm$2.28 & \underline{96.79$\pm$0.26} & 82.49$\pm$3.32 & 60.37$\pm$1.31 & 85.97 \\
T3A \cite{T3A} & 98.89$\pm$1.57 & \textbf{95.31$\pm$0.14} & 96.22$\pm$0.16 & 77.20$\pm$2.80 & 59.40$\pm$1.96 & 85.40 \\
TTAC \cite{TTAC} & 98.89$\pm$1.57 & 94.04$\pm$1.37 & 95.79$\pm$0.26 & 76.61$\pm$0.48 & 59.42$\pm$2.04 & 84.95 \\
TAST \cite{TAST} & \textbf{99.55$\pm$0.42} & 95.31$\pm$0.14 & 96.32$\pm$0.31 & \textbf{85.84$\pm$1.23} & 60.81$\pm$2.33 & \underline{87.57} \\
RoTTA \cite{RoTTA} & 98.66$\pm$1.89 & 93.21$\pm$0.73 & 95.59$\pm$0.39 & 79.92$\pm$2.32 & 59.10$\pm$1.32 & 85.29 \\
TSD \cite{TSD} & 98.67$\pm$0.98 & 92.08$\pm$2.03 & 96.24$\pm$0.30 & 81.87$\pm$3.20 & 60.26$\pm$1.22 & 85.82 \\
\midrule
\textbf{Proposed}  & 98.66$\pm$1.42  & 95.13$\pm$0.13  & \textbf{98.94$\pm$0.40}  & 83.45$\pm$3.39  & \textbf{64.60$\pm$4.92}  & \textbf{88.16} \\ 
 \bottomrule
\end{NiceTabular}}
\label{ucihar_main_results}
\end{table*}




\begin{table*}[h]
\centering
\caption{Detailed results of the five MFD cross-domain scenarios in terms of macro F1 score.}
\setlength{\tabcolsep}{2.5mm}{
\begin{NiceTabular}{@{}l|ccccc|c@{}} 
\toprule 
Algorithm & 0$\rightarrow$1 & 1$\rightarrow$2 & 3$\rightarrow$1 & 1$\rightarrow$0 & 2$\rightarrow$3 & Average\\ \midrule
Target & 99.89$\pm$0.19 & 99.95$\pm$0.05 & 99.89$\pm$0.19 & 98.78$\pm$1.84 & 100.0$\pm$0.00 & 99.70 \\
\midrule
Source & 48.08$\pm$3.82 & 78.45$\pm$2.12 & \underline{99.98$\pm$0.03} & 72.68$\pm$2.93 & 99.53$\pm$0.24 & 79.74 \\
BN \cite{BN} & 97.88$\pm$0.68 & 88.05$\pm$0.92 & \underline{99.98$\pm$0.01} & 88.32$\pm$0.73 & 98.32$\pm$0.20 & 94.51 \\
TENT \cite{TENT} & 97.93$\pm$0.71 & 88.02$\pm$0.90 & \underline{99.98$\pm$0.01} & 88.35$\pm$0.68 & 98.31$\pm$0.19 & 94.52 \\
PL \cite{PL} & \underline{98.77$\pm$0.25} & 87.42$\pm$1.09 & \underline{99.98$\pm$0.01} & \underline{88.57$\pm$0.57} & 98.43$\pm$0.24 & \underline{94.64}  \\
SHOT-IM \cite{SHOT_IM} & 75.22$\pm$4.51 & 86.17$\pm$2.94 & \textbf{100.0$\pm$0.00} & 67.87$\pm$5.70 & \textbf{99.84$\pm$0.08} & 85.82  \\
EATA \cite{EATA} & 98.20$\pm$0.45 & 87.93$\pm$0.92 & \underline{99.98$\pm$0.01} & 88.30$\pm$0.54 & 98.30$\pm$0.22 & 94.54 \\
NOTE \cite{NOTE} & 97.92$\pm$0.71 & 88.03$\pm$0.91 & \underline{99.98$\pm$0.01} & 88.32$\pm$0.72 & 98.32$\pm$0.20 & 94.51 \\
CoTTA \cite{CoTTA} & 97.58$\pm$1.28 & 87.63$\pm$0.65 & \underline{99.98$\pm$0.01} & 88.41$\pm$0.28 & 98.41$\pm$0.13 & 94.40 \\
SAR \cite{SAR} & 97.92$\pm$0.71 & \underline{88.06$\pm$0.92} & \underline{99.98$\pm$0.01} & 88.32$\pm$0.73 & 98.32$\pm$0.20 & 94.52 \\
T3A \cite{T3A} & 54.70$\pm$12.36 & 79.89$\pm$5.13 & \textbf{100.0$\pm$0.00} & 63.65$\pm$4.96 & 97.68$\pm$0.72 & 79.18 \\
TTAC \cite{TTAC} & 91.49$\pm$2.42 & 85.68$\pm$1.79 & \underline{99.98$\pm$0.03} & 86.71$\pm$0.66 & 96.83$\pm$1.67 & 92.14 \\
TAST \cite{TAST} & 95.14$\pm$6.78 & 83.41$\pm$5.76 & \textbf{100.0$\pm$0.00} & 70.04$\pm$5.16 & 95.44$\pm$3.07 & 88.81 \\
RoTTA \cite{RoTTA} & 86.57$\pm$4.58 & 85.83$\pm$1.84 & \textbf{100.0$\pm$0.00} & 81.10$\pm$4.33 & 99.72$\pm$0.11 & 90.64 \\
TSD \cite{TSD} & 97.87$\pm$0.73 & 88.05$\pm$0.90 & \underline{99.98$\pm$0.01} & 88.31$\pm$0.74 & 98.32$\pm$0.20 & 94.51 \\
\midrule
\textbf{Proposed}  & \textbf{99.64$\pm$0.10} & \textbf{89.69$\pm$0.53} & \textbf{100.0$\pm$0.00} & \textbf{88.87$\pm$5.16} & \underline{99.78$\pm$0.03} & \textbf{95.60} \\ 
 \bottomrule
\end{NiceTabular}}
\label{mfd_main_results}
\end{table*}





\begin{table*}[h]
\centering
\caption{Detailed results of the five SSC cross-domain scenarios in terms of macro F1 score.}
\setlength{\tabcolsep}{2.5mm}{
\begin{NiceTabular}{@{}l|ccccc|c@{}} 
\toprule 
Algorithm & 0$\rightarrow$11 & 12$\rightarrow$5 & 7$\rightarrow$18 & 16$\rightarrow$1 & 9$\rightarrow$14 & Average\\ \midrule
Target & 65.73$\pm$2.08 & 75.19$\pm$3.61 & 73.97$\pm$1.87 & 78.03$\pm$2.53 & 75.57$\pm$1.17 & 73.70 \\
\midrule
Source & 39.85$\pm$3.19 & 53.59$\pm$0.75 & 57.63$\pm$2.24 & 42.18$\pm$1.88 & 58.67$\pm$3.60 & 50.38 \\
BN \cite{BN} & 43.29$\pm$3.15 & 57.57$\pm$1.56 & 68.42$\pm$1.18 & 58.81$\pm$1.26 & 63.31$\pm$1.76 & 58.28 \\
TENT \cite{TENT} & 43.35$\pm$3.04 & 57.66$\pm$1.59 & 68.46$\pm$1.13 & 58.89$\pm$1.22 & 63.43$\pm$1.83 & 58.36 \\
PL \cite{PL} & 43.33$\pm$3.09 & 57.69$\pm$1.54 & 68.41$\pm$1.11 & 58.88$\pm$1.25 & 63.52$\pm$1.75 & 58.37  \\
SHOT-IM \cite{SHOT_IM} & 41.49$\pm$2.51 & 60.00$\pm$1.11 & 65.98$\pm$1.24 & 58.43$\pm$0.50 & 66.66$\pm$2.30 & 58.51  \\
EATA \cite{EATA} & 43.36$\pm$3.35 & 58.38$\pm$1.42 & 68.37$\pm$1.28 & 58.73$\pm$1.31 & 64.07$\pm$1.46 & 58.58 \\
NOTE \cite{NOTE} & 43.32$\pm$3.13 & 57.77$\pm$1.52 & 68.48$\pm$1.13 & 58.82$\pm$1.27 & 63.48$\pm$1.72 & 58.37 \\
CoTTA \cite{CoTTA} & 42.47$\pm$3.95 & 57.40$\pm$1.61 & 67.84$\pm$1.12 & 59.24$\pm$0.76 & 63.88$\pm$1.53 & 58.17 \\
SAR \cite{SAR} & 43.27$\pm$3.03 & 57.74$\pm$1.56 & 68.47$\pm$1.15 & 58.84$\pm$1.18 & 63.42$\pm$1.87 & 58.35 \\
T3A \cite{T3A} & 42.68$\pm$4.43 & \textbf{64.70$\pm$2.47} & 66.41$\pm$2.28 & 55.64$\pm$1.29 & \underline{69.56$\pm$2.38} & \underline{59.80} \\
TTAC \cite{TTAC} & 43.07$\pm$3.83 & 55.47$\pm$0.58 & 65.83$\pm$0.52 & 56.30$\pm$0.66 & 59.17$\pm$4.38 & 55.97 \\
TAST \cite{TAST} & 38.32$\pm$6.93 & 59.14$\pm$5.16 & \underline{70.95$\pm$1.03} & 57.56$\pm$1.69 & 64.82$\pm$3.52 & 58.16 \\
RoTTA \cite{RoTTA} & \underline{44.67$\pm$4.16} & 51.64$\pm$0.87 & 68.96$\pm$1.06 & \underline{60.11$\pm$0.73} & 56.31$\pm$2.76 & 56.34 \\
TSD \cite{TSD} & 43.28$\pm$3.12 & 57.82$\pm$1.46 & 68.38$\pm$1.17 & 58.76$\pm$1.28 & 63.42$\pm$1.84 & 58.33 \\
\midrule
\textbf{Proposed}  & \textbf{45.10$\pm$5.49} & \underline{63.99$\pm$1.57} & \textbf{71.77$\pm$0.68} & \textbf{62.35$\pm$1.08} & \textbf{70.02$\pm$1.17} & \textbf{62.65} \\ 
 \bottomrule
\end{NiceTabular}}
\label{ssc_main_results}
\end{table*}





We evaluate the model performance on three commonly used time series datasets from three real-world scenarios, i.e., human activity recognition (UCIHAR) \cite{har_dataset}, machine fault diagnosis (MFD) \cite{mfd_dataset}, and sleep stage classification (SSC) \cite{ssc_dataset}. The details of each dataset are presented in the Appendix \ref{intro_datasets}.

\subsection{Baseline Methods}
For evaluation, we compare our method with following established test-time adaptation approaches: BN \cite{BN}, PL \cite{PL}, TENT ~\cite{TENT}, SHOT \cite{SHOT_IM}, EATA ~\cite{EATA}, SAR \cite{SAR}, TTAC \cite{TTAC}, NOTE \cite{NOTE}, T3A \cite{T3A}, TAST \cite{TAST}, CoTTA \cite{CoTTA}, RoTTA \cite{RoTTA}, and TSD \cite{TSD}. Additionally, we include the method of source-only (i.e., Source) and target-supervised (i.e., Target) performance to establish baseline and upper bound on generalizability, respectively. The description of each comparison method is shown in Appendix \ref{intro_baseline_methods}.

\subsection{Comparative Experiments}
We evaluate our model's efficacy on three real-world time-series datasets: UCIHAR, MFD, and SSC. The experimental results of the different approaches for five cross-domain scenarios for each dataset, as well as the average results across all the scenarios, are presented in Tables \ref{ucihar_main_results}, \ref{mfd_main_results} and \ref{ssc_main_results}, respectively. In addition, to further validate the versatility potential of the proposed model, we also conduct an additional comparative experiment on the visual dataset PACS \cite{PACS}. Detailed results are provided in Appendix \ref{appendix_generalization_vision_tasks}.

\subsubsection{Evaluation on UCIHAR Dataset}
Table \ref{ucihar_main_results} presents the classification performance of various methods on the UCIHAR dataset across five cross-subject scenarios. The results indicate performance improvements for all methods when compared to the direct testing using the source domain pre-trained model. Methods trained using entropy minimization or pseudo-labeling (e.g. TENT, PL, SHOT, EATA, SAR) all achieve comparable classification performance, while the TAST method further enhances prediction reliability by introducing prototype prediction information from nearest neighbors. Distinctly, our method excelled in two scenarios and achieved an overall macro F1 score of 88.16\%. This success likely stems from leveraging augmented contrastive clustering to effectively group similar samples while differentiating dissimilar ones. Moreover, the uncertainty-aware prototypical ensemble further enhances the quality of pseudo-labels.

\subsubsection{Evaluation on MFD Dataset}
Table \ref{mfd_main_results} showcases our method's dominance on the MFD task, achieving an average macro F1 score of 95.60\% outperforming other methods in four scenarios. It significantly improves upon the Source method by over 15\% and surpasses the second-best approach by nearly 1\%. Notably, most methods on MFD see substantial performance gains, likely due to the longer sequences offering richer information and the smaller number of categories facilitating improvement. Nevertheless, T3A and TAST exhibit relatively lower performance, possibly due to substantial distributional differences between the source and target domains in this dataset. The use of prototype classifications alone may not perfectly characterize the distributional properties of the target domains, resulting in marginal performance gains.

\subsubsection{Evaluation on SSC Dataset}
Table \ref{ssc_main_results} highlights our method's superiority on the sleep staging task, achieving an average macro F1 score of 62.65\% and excelling in four out of five scenarios. This outperforms the second-best method by 2.85\% and improves upon the source pre-trained model by 12.27\%, respectively. This can be attributed to the inherent characteristic of class imbalance in the SSC dataset, which tends to limit the performance improvement of other methods. In contrast, our method mitigates the degradation by introducing an uncertainty-aware prototype ensemble and augmented contrastive clustering, resulting in significantly enhanced adaptation performance.

\subsection{Ablation Studies}
In this section, we perform various ablation studies to assess the individual contributions within each module on three datasets. In addition, we also provide overall ablation study to quantify the effect of each module of proposed model in Appendix \ref{appendix_ablation_of_each_module}.

\subsubsection{Effectiveness of Uncertainty-Aware Prototypical Ensemble}
To explore the impact of our uncertainty-aware prototypical ensemble, we evaluated different pseudo-labeling strategies, as shown in Table \ref{different_pseudo_labels}.  Specifically, $p_{ensemble}$ denotes pseudo-labels predicted with augmentation-ensemble. $p_{prototypes.}$ denotes pseudo-labels obtained with uncertainty-aware prototypes. $p_{averaged}$ denotes pseudo-labels obtained with averaged predictions using augmentation-ensemble and uncertainty-aware prototypes. While averaging prototypes and enhanced predictions can achieve competent performance, it may harm performance on certain datasets, leading to a reduction of around 3\% due to its disregard for prediction uncertainty during averaging. In contrast, our method selects more reliable predictions rather than averaging both. Besides, our method outperformed other pseudo-labeling strategies on two of the three datasets and achieved the second-highest performance on the third. This demonstrates that our entropy comparison scheme significantly enhances the quality of pseudo-labels, resulting in superior outcomes compared to alternative approaches. This also suggested that, in general, data with longer sequences and fewer categories tend to benefit from classification-based logits due to richer discriminative information. Conversely, datasets with a larger number of classes and imbalanced class distributions may favour prototype-based prediction.

\begin{table}[h]
\centering
\caption{Ablation study of different pseudo-labels strategies.}
\setlength{\tabcolsep}{4.5mm}{
\begin{NiceTabular}{@{}l|ccc@{}} 
\toprule 
Variants & UCIHAR & MFD & SSC \\ \midrule
$p_{ensemble}$ & 86.82 & \textbf{95.64} & 58.36   \\
$p_{prototypes}$ & 87.24 & 94.42  & 58.67  \\ 
$p_{averaged}$ & 87.92 & 92.65 & 62.64   \\ \midrule
\textbf{Proposed} & \textbf{88.16} & 95.60 & \textbf{62.65}  \\
\bottomrule
\end{NiceTabular}}
\label{different_pseudo_labels}
\end{table}


\begin{figure*}[h]
\centering
\begin{tabular}{cc}
\subfigure[]{\includegraphics[width=0.33\linewidth]{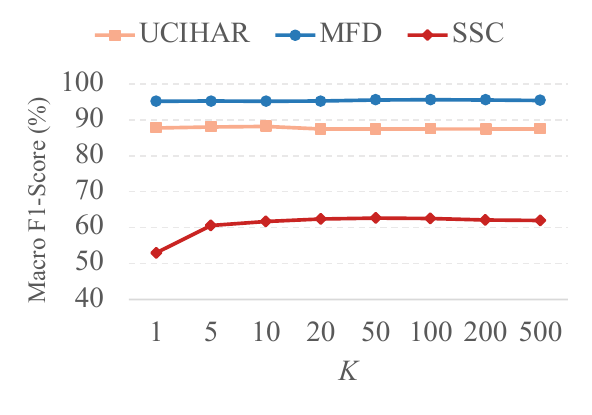} \label{figure4_a}}
\subfigure[]{\includegraphics[width=0.33\linewidth]{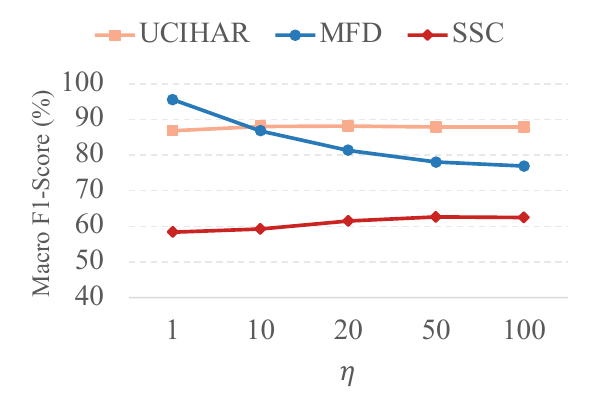} \label{figure4_b}}
\subfigure[]{\includegraphics[width=0.33\linewidth]{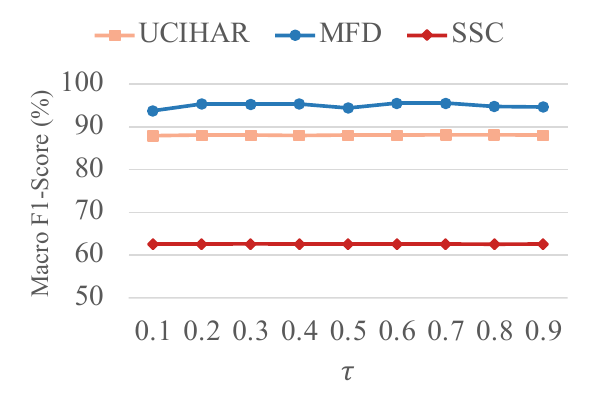} \label{figure4_c}}
\end{tabular}
\caption{Analysis of adaptation performance with varying parameters. (a) $K$ for the memorized support set selection. (b) $\eta$ for uncertainty-aware prototypes component. (c) $\tau$ for augmented contrastive clustering component.}
\label{different_optimization_objs}
\end{figure*}

\subsubsection{Effectiveness of Augmented Contrastive Clustering}
We validate the effectiveness of augmented contrastive clustering through ablation experiments using various optimization objectives, as outlined in Table \ref{tab_different_optimization_objs}. Specifically, \textit{SE Loss} refers to entropy minimization as the optimization objective, \textit{CE Loss} indicates the use of cross-entropy loss, while \textit{TC Loss} denotes the use of traditional contrastive laerning. Our augmented contrastive clustering objective consistently outperforms the others across all three datasets, highlighting its efficacy in mitigating error accumulation even in the presence of a few noisy labels. In addition, in comparison to traditional contrastive learning, our approach, by focusing on cross-cluster contrasts, yields more robust representations that exhibit greater resilience to noisy label conditions.

\begin{table}[h]
\centering
\caption{Ablation study of different optimization objectives.}
\setlength{\tabcolsep}{4.5mm}{
\begin{NiceTabular}{@{}l|ccc@{}} 
\toprule 
Variants & UCIHAR & MFD & SSC \\ \midrule
SE Loss & 87.34 & 93.04 & 62.54  \\
CE Loss & 87.20 & 95.13 & 62.64  \\ 
TC Loss & 87.46 & 94.72 & 61.64  \\ 
\midrule
\textbf{Proposed} & \textbf{88.16} & \textbf{95.60} & \textbf{62.65}  \\
\bottomrule
\end{NiceTabular}}
\label{tab_different_optimization_objs}
\end{table}


\subsubsection{Effectiveness of Different Augmentation Strategies}
We investigated the impact of various data augmentation strategies on performance. Table \ref{different_augmentations} compares our employed magnitude warping augmentation method with other commonly used techniques \cite{magnitude_warp_aug}. Time series specific magnitude warping augmentation achieves the best performance across two out of three datasets. This success may be attributed to the elastic transformations introduced by this method, which preserve the inherent temporal dynamics of the time series data while introducing nuanced variations. Additionally, all augmentation methods resulted in improved classification performance compared to no augmentation, underscoring the effectiveness of employing appropriate weak augmentation strategies to enhance model features and bolster prediction robustness. We also conduct additional experiments to explore the effects of different augmentation combinations in the Appendix \ref{appendix_different_augmentation_combination}.

\begin{table}[h]
\centering
\caption{Ablation experiments using different augmentation strategies.}
\setlength{\tabcolsep}{4.5mm}{
\begin{NiceTabular}{@{}l|ccc@{}} 
\toprule 
Augmentations & UCIHAR & MFD & SSC \\ \midrule
no Aug & 87.79 & 95.10 & 62.47 \\
Jitter & 87.93 & 95.16 & 62.49 \\
Scale & 87.84 & 95.59 & 62.66 \\
Permutation & 81.78 & 95.15 & \textbf{63.15} \\ \midrule
\textbf{Proposed} & \textbf{88.16} & \textbf{95.60} & 62.65 \\
\bottomrule
\end{NiceTabular}}
\label{different_augmentations}
\end{table}


\subsection{Model Analysis}
Here, we perform parameter sensitivity analysis and qualitative analysis. We also analyze the model performance when fine-tuning different layers of the feature encoder, as shown in Appendix \ref{fine_tune_different_layers}.

\subsubsection{Parameter Sensitivity Analysis}
Our model depends on three key parameters: $K$ for support set selection, $\eta$ for prototype prediction tuning, and $\tau$ for contrastive clustering adjustment. Sensitivity analyses of these parameters on different datasets are shown in Figures \ref{figure4_a}- \ref{figure4_c}. We consider a candidate parameter set: $K \in \left\{1, 5, 10, 20, 50, 100, 200, 500\right\}$, $\eta \in \left\{1, 10, 20, 50, 100\right\}$, and $\tau \in \left[0.1, 0.9\right]$ with a $0.1$ interval. For $K$, values in $\left\{5, 10\right\}$ work best for UCIHAR, while $\left\{50, 100\right\}$ are superior for MFD and SSC. Regarding $\eta$, smaller values are optimal for MFD, larger for SSC, with UCIHAR showing insensitivity. For $\tau$, the model performs best at $0.7$, $0.6$, and $0.3$ on UCIHAR, MFD, and SSC, respectively.

\subsubsection{Qualitative Analysis by t-SNE}
\begin{figure}[t]
\centering
\begin{tabular}{cc}
\subfigure[]{\includegraphics[width=0.33\linewidth]{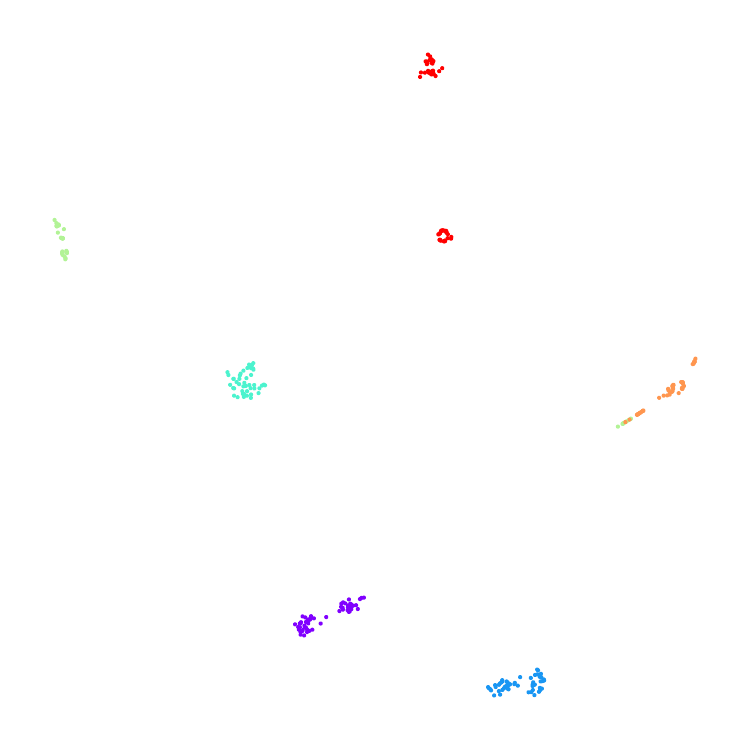} \label{figure5_a}}
\subfigure[]{\includegraphics[width=0.33\linewidth]{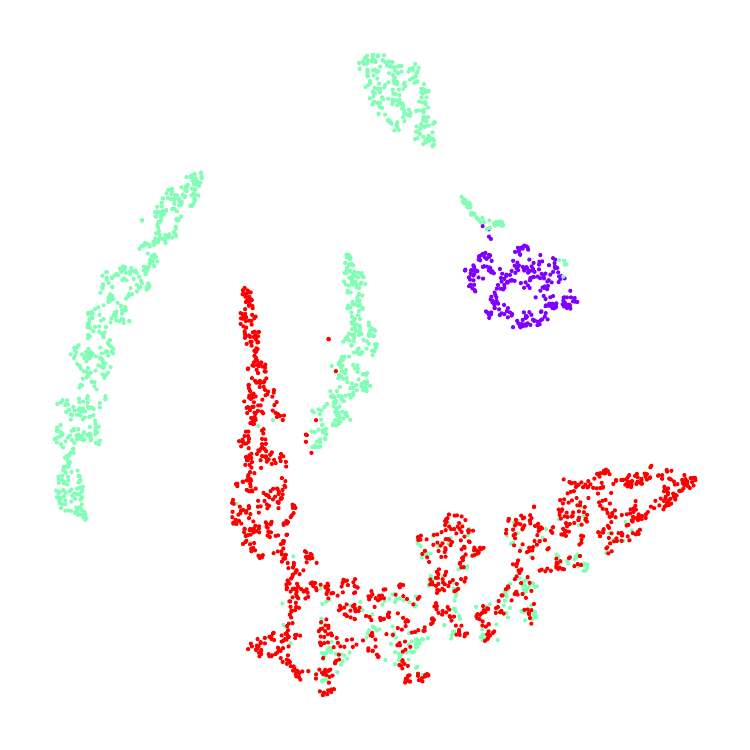} \label{figure5_b}}
\subfigure[]{\includegraphics[width=0.33\linewidth]{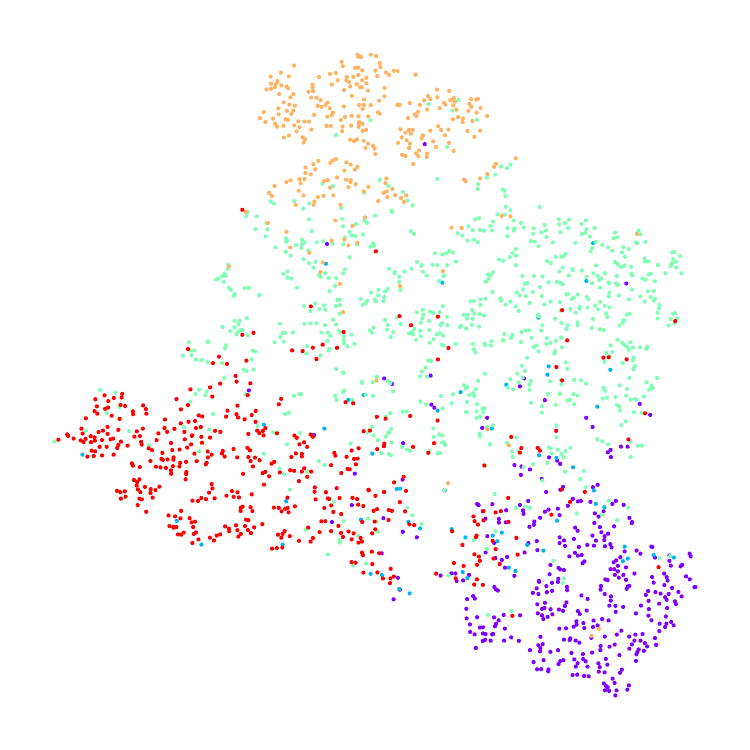} \label{figure5_c}} \\
\subfigure[]{\includegraphics[width=0.33\linewidth]{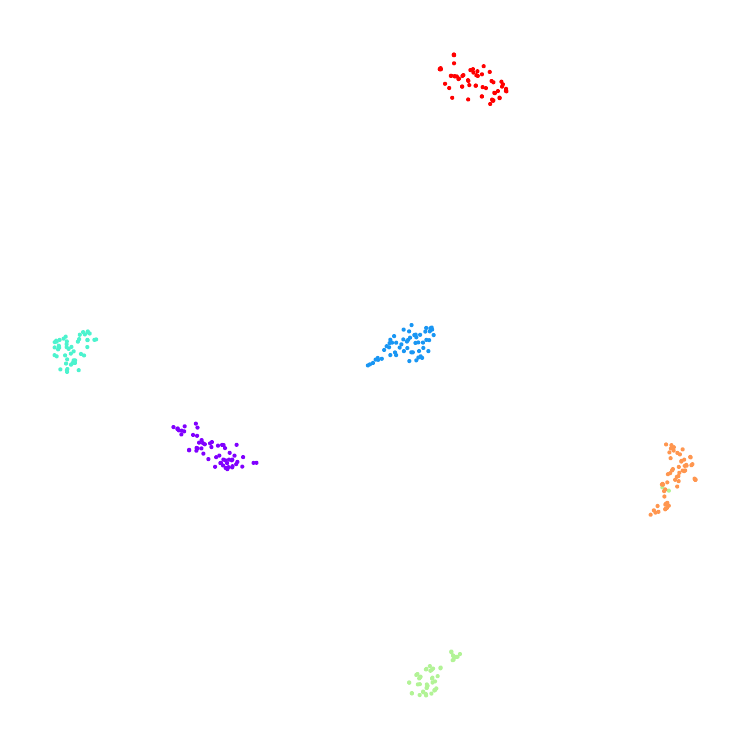} \label{figure5_d}} 
\subfigure[]{\includegraphics[width=0.33\linewidth]{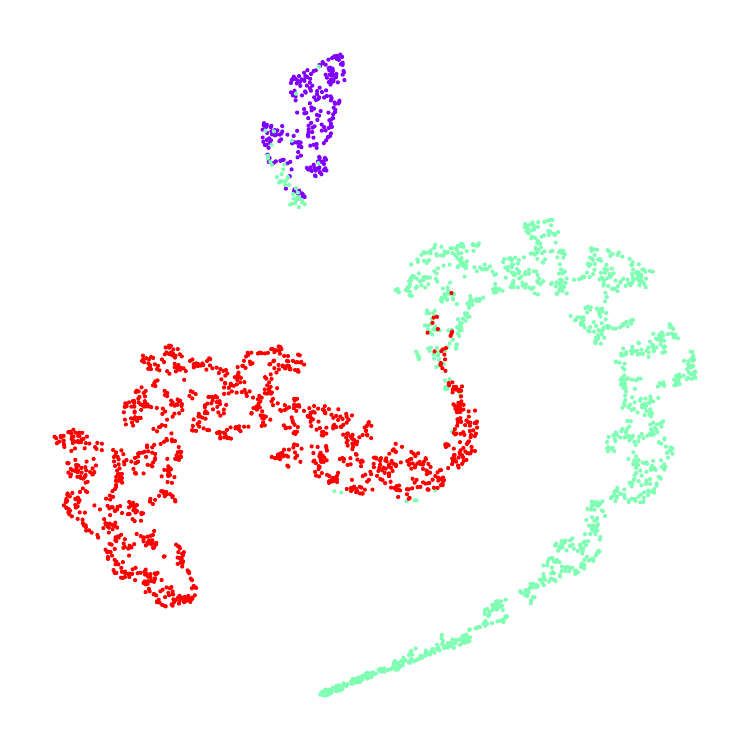} \label{figure5_e}}
\subfigure[]{\includegraphics[width=0.33\linewidth]{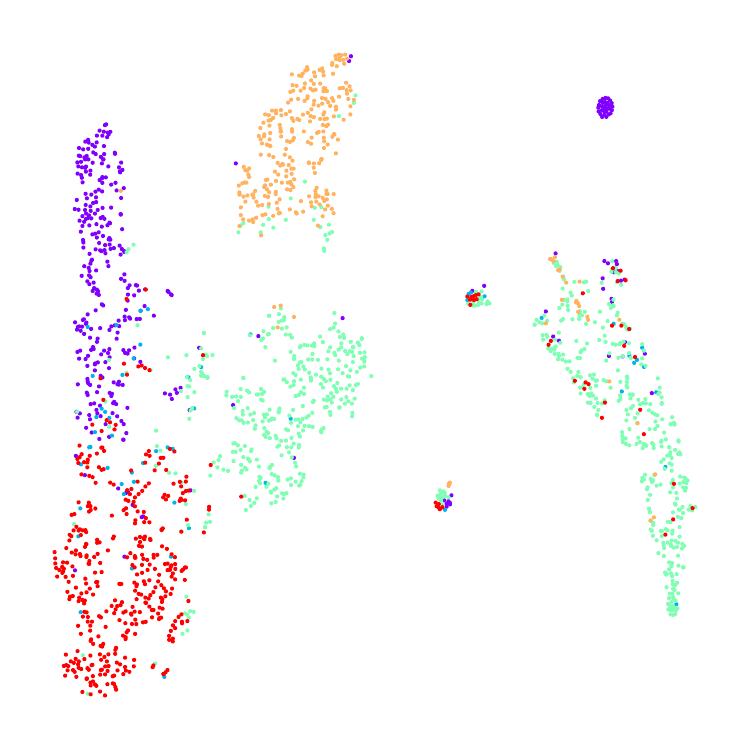} \label{figure5_f}} 
\end{tabular}
\caption{t-SNE visualization of extracted feature distributions. Colors denote different categories. (a) depicts the source pre-trained model without adaptation on the UCIHAR dataset, (b) MFD dataset, and (c) SSC dataset. Following adaptation with the proposed model, (d) shows the MFD dataset, (e) UCIHAR dataset, and (f) SSC dataset.}
\label{figure5}
\end{figure}

We present qualitative results for test-time adaptation by visualizing feature distributions of the source baseline and after adaptation using proposed method via t-SNE on various datasets. In figure \ref{figure5_a}-\ref{figure5_c}, it is evident that the features learned by the source baseline do not separate well due to large domain distribution discrepancies and exhibit the same category of features that do not aggregate well in a cluster. Upon adapting the model to the target domain data using our method, as depicted in Figure \ref{figure5_d}-\ref{figure5_e}, the extracted features demonstrate improved separation, with features of the same class predominantly clustered together. This demonstrates the effectiveness of augmented contrastive clustering in encouraging cohesive clustering within categories and promoting clear separation between them.

\section{Conclusion}
This paper introduced augmented contrastive clustering with uncertainty aware prototyping (ACCUP), a novel test-time adaptation method specifically designed for real-world time series data. ACCUP tackles the challenge of unreliable pseudo-labels and unsatisfactory model adaptation due to the noisy and non-stationary characteristics of real-world time series through uncertainty-aware prototypical ensemble and augmented contrastive clustering strategies. This is the first method dedicated to test-time adaptation in generic real-world time series applications. Extensive experiments on four diverse datasets demonstrate the validity of the proposed method, achieving state-of-the-art performance compared to existing methods. This work highlights the potential of deploying models in time series applications to address domain shifting in more realistic environments while preserving data privacy.

\begin{acks}
This work was supported by the National Natural Science Foundation of China (Nos. 62136004, 62276130), by the National Key R\&D Program of China (No. 2023YFF1204803), and also by the National Research Foundation, Singapore under its AI Singapore Programme (AISG2-RP-2021-027).
\end{acks}

\bibliographystyle{ACM-Reference-Format}
\bibliography{ref}


\begin{thebibliography}{72}


\ifx \showCODEN    \undefined \def \showCODEN     #1{\unskip}     \fi
\ifx \showDOI      \undefined \def \showDOI       #1{#1}\fi
\ifx \showISBNx    \undefined \def \showISBNx     #1{\unskip}     \fi
\ifx \showISBNxiii \undefined \def \showISBNxiii  #1{\unskip}     \fi
\ifx \showISSN     \undefined \def \showISSN      #1{\unskip}     \fi
\ifx \showLCCN     \undefined \def \showLCCN      #1{\unskip}     \fi
\ifx \shownote     \undefined \def \shownote      #1{#1}          \fi
\ifx \showarticletitle \undefined \def \showarticletitle #1{#1}   \fi
\ifx \showURL      \undefined \def \showURL       {\relax}        \fi
\providecommand\bibfield[2]{#2}
\providecommand\bibinfo[2]{#2}
\providecommand\natexlab[1]{#1}
\providecommand\showeprint[2][]{arXiv:#2}

\bibitem[Anguita et~al\mbox{.}(2013)]%
        {har_dataset}
\bibfield{author}{\bibinfo{person}{Davide Anguita}, \bibinfo{person}{Alessandro Ghio}, \bibinfo{person}{Luca Oneto}, \bibinfo{person}{Xavier Parra}, \bibinfo{person}{Jorge~Luis Reyes-Ortiz}, {et~al\mbox{.}}} \bibinfo{year}{2013}\natexlab{}.
\newblock \showarticletitle{A public domain dataset for human activity recognition using smartphones.}. In \bibinfo{booktitle}{\emph{Proceedings of the European Symposium on Artificial Neural Networks}}, Vol.~\bibinfo{volume}{3}. \bibinfo{pages}{437–442}.
\newblock


\bibitem[Boudiaf et~al\mbox{.}(2022)]%
        {LAME}
\bibfield{author}{\bibinfo{person}{Malik Boudiaf}, \bibinfo{person}{Romain Mueller}, \bibinfo{person}{Ismail Ben~Ayed}, {and} \bibinfo{person}{Luca Bertinetto}.} \bibinfo{year}{2022}\natexlab{}.
\newblock \showarticletitle{Parameter-free online test-time adaptation}. In \bibinfo{booktitle}{\emph{Proceedings of the IEEE/CVF Conference on Computer Vision and Pattern Recognition}}. \bibinfo{pages}{8344--8353}.
\newblock


\bibitem[Brahma and Rai(2023)]%
        {PETAL}
\bibfield{author}{\bibinfo{person}{Dhanajit Brahma} {and} \bibinfo{person}{Piyush Rai}.} \bibinfo{year}{2023}\natexlab{}.
\newblock \showarticletitle{A probabilistic framework for lifelong test-time adaptation}. In \bibinfo{booktitle}{\emph{Proceedings of the IEEE/CVF Conference on Computer Vision and Pattern Recognition}}. \bibinfo{pages}{3582--3591}.
\newblock


\bibitem[Cai et~al\mbox{.}(2021)]%
        {UDA_TS1}
\bibfield{author}{\bibinfo{person}{Ruichu Cai}, \bibinfo{person}{Jiawei Chen}, \bibinfo{person}{Zijian Li}, \bibinfo{person}{Wei Chen}, \bibinfo{person}{Keli Zhang}, \bibinfo{person}{Junjian Ye}, \bibinfo{person}{Zhuozhang Li}, \bibinfo{person}{Xiaoyan Yang}, {and} \bibinfo{person}{Zhenjie Zhang}.} \bibinfo{year}{2021}\natexlab{}.
\newblock \showarticletitle{Time series domain adaptation via sparse associative structure alignment}. In \bibinfo{booktitle}{\emph{Proceedings of the AAAI Conference on Artificial Intelligence}}, Vol.~\bibinfo{volume}{35}. \bibinfo{pages}{6859--6867}.
\newblock


\bibitem[Chen et~al\mbox{.}(2019)]%
        {prototype_Int1}
\bibfield{author}{\bibinfo{person}{Chaofan Chen}, \bibinfo{person}{Oscar Li}, \bibinfo{person}{Daniel Tao}, \bibinfo{person}{Alina Barnett}, \bibinfo{person}{Cynthia Rudin}, {and} \bibinfo{person}{Jonathan~K Su}.} \bibinfo{year}{2019}\natexlab{}.
\newblock \showarticletitle{This looks like that: deep learning for interpretable image recognition}.
\newblock \bibinfo{journal}{\emph{Advances in neural information processing systems}}  \bibinfo{volume}{32} (\bibinfo{year}{2019}).
\newblock


\bibitem[Chen et~al\mbox{.}(2022)]%
        {AdaContrast}
\bibfield{author}{\bibinfo{person}{Dian Chen}, \bibinfo{person}{Dequan Wang}, \bibinfo{person}{Trevor Darrell}, {and} \bibinfo{person}{Sayna Ebrahimi}.} \bibinfo{year}{2022}\natexlab{}.
\newblock \showarticletitle{Contrastive test-time adaptation}. In \bibinfo{booktitle}{\emph{Proceedings of the IEEE/CVF Conference on Computer Vision and Pattern Recognition}}. \bibinfo{pages}{295--305}.
\newblock


\bibitem[Cover and Hart(1967)]%
        {prototype_early}
\bibfield{author}{\bibinfo{person}{Thomas Cover} {and} \bibinfo{person}{Peter Hart}.} \bibinfo{year}{1967}\natexlab{}.
\newblock \showarticletitle{Nearest neighbor pattern classification}.
\newblock \bibinfo{journal}{\emph{IEEE transactions on information theory}} \bibinfo{volume}{13}, \bibinfo{number}{1} (\bibinfo{year}{1967}), \bibinfo{pages}{21--27}.
\newblock


\bibitem[De~Lange and Tuytelaars(2021)]%
        {prototype_CL2}
\bibfield{author}{\bibinfo{person}{Matthias De~Lange} {and} \bibinfo{person}{Tinne Tuytelaars}.} \bibinfo{year}{2021}\natexlab{}.
\newblock \showarticletitle{Continual prototype evolution: Learning online from non-stationary data streams}. In \bibinfo{booktitle}{\emph{Proceedings of the IEEE/CVF international conference on computer vision}}. \bibinfo{pages}{8250--8259}.
\newblock


\bibitem[Eldele et~al\mbox{.}(2023)]%
        {CA-TCC}
\bibfield{author}{\bibinfo{person}{Emadeldeen Eldele}, \bibinfo{person}{Mohamed Ragab}, \bibinfo{person}{Zhenghua Chen}, \bibinfo{person}{Min Wu}, \bibinfo{person}{Chee-Keong Kwoh}, \bibinfo{person}{Xiaoli Li}, {and} \bibinfo{person}{Cuntai Guan}.} \bibinfo{year}{2023}\natexlab{}.
\newblock \showarticletitle{Self-supervised contrastive representation learning for semi-supervised time-series classification}.
\newblock \bibinfo{journal}{\emph{IEEE Transactions on Pattern Analysis and Machine Intelligence}} (\bibinfo{year}{2023}).
\newblock


\bibitem[Elsayed et~al\mbox{.}(2018)]%
        {cross_entropy_reason3}
\bibfield{author}{\bibinfo{person}{Gamaleldin Elsayed}, \bibinfo{person}{Dilip Krishnan}, \bibinfo{person}{Hossein Mobahi}, \bibinfo{person}{Kevin Regan}, {and} \bibinfo{person}{Samy Bengio}.} \bibinfo{year}{2018}\natexlab{}.
\newblock \showarticletitle{Large margin deep networks for classification}.
\newblock \bibinfo{journal}{\emph{Advances in neural information processing systems}}  \bibinfo{volume}{31} (\bibinfo{year}{2018}).
\newblock


\bibitem[Goldberger et~al\mbox{.}(2000)]%
        {ssc_dataset}
\bibfield{author}{\bibinfo{person}{Ary~L Goldberger}, \bibinfo{person}{Luis~AN Amaral}, \bibinfo{person}{Leon Glass}, \bibinfo{person}{Jeffrey~M Hausdorff}, \bibinfo{person}{Plamen~Ch Ivanov}, \bibinfo{person}{Roger~G Mark}, \bibinfo{person}{Joseph~E Mietus}, \bibinfo{person}{George~B Moody}, \bibinfo{person}{Chung-Kang Peng}, {and} \bibinfo{person}{H~Eugene Stanley}.} \bibinfo{year}{2000}\natexlab{}.
\newblock \showarticletitle{PhysioBank, PhysioToolkit, and PhysioNet: components of a new research resource for complex physiologic signals}.
\newblock \bibinfo{journal}{\emph{circulation}} \bibinfo{volume}{101}, \bibinfo{number}{23} (\bibinfo{year}{2000}), \bibinfo{pages}{e215--e220}.
\newblock


\bibitem[Gong et~al\mbox{.}(2022)]%
        {NOTE}
\bibfield{author}{\bibinfo{person}{Taesik Gong}, \bibinfo{person}{Jongheon Jeong}, \bibinfo{person}{Taewon Kim}, \bibinfo{person}{Yewon Kim}, \bibinfo{person}{Jinwoo Shin}, {and} \bibinfo{person}{Sung-Ju Lee}.} \bibinfo{year}{2022}\natexlab{}.
\newblock \showarticletitle{Note: Robust continual test-time adaptation against temporal correlation}.
\newblock \bibinfo{journal}{\emph{Advances in Neural Information Processing Systems}}  \bibinfo{volume}{35} (\bibinfo{year}{2022}), \bibinfo{pages}{27253--27266}.
\newblock


\bibitem[Gong et~al\mbox{.}(2024)]%
        {SoTTA}
\bibfield{author}{\bibinfo{person}{Taesik Gong}, \bibinfo{person}{Yewon Kim}, \bibinfo{person}{Taeckyung Lee}, \bibinfo{person}{Sorn Chottananurak}, {and} \bibinfo{person}{Sung-Ju Lee}.} \bibinfo{year}{2024}\natexlab{}.
\newblock \showarticletitle{SoTTA: Robust Test-Time Adaptation on Noisy Data Streams}.
\newblock \bibinfo{journal}{\emph{Advances in Neural Information Processing Systems}}  \bibinfo{volume}{36} (\bibinfo{year}{2024}).
\newblock


\bibitem[Goyal et~al\mbox{.}(2022)]%
        {conjugateTTA}
\bibfield{author}{\bibinfo{person}{Sachin Goyal}, \bibinfo{person}{Mingjie Sun}, \bibinfo{person}{Aditi Raghunathan}, {and} \bibinfo{person}{J~Zico Kolter}.} \bibinfo{year}{2022}\natexlab{}.
\newblock \showarticletitle{Test time adaptation via conjugate pseudo-labels}.
\newblock \bibinfo{journal}{\emph{Advances in Neural Information Processing Systems}}  \bibinfo{volume}{35} (\bibinfo{year}{2022}), \bibinfo{pages}{6204--6218}.
\newblock


\bibitem[He et~al\mbox{.}(2023)]%
        {UDA_TS6}
\bibfield{author}{\bibinfo{person}{Huan He}, \bibinfo{person}{Owen Queen}, \bibinfo{person}{Teddy Koker}, \bibinfo{person}{Consuelo Cuevas}, \bibinfo{person}{Theodoros Tsiligkaridis}, {and} \bibinfo{person}{Marinka Zitnik}.} \bibinfo{year}{2023}\natexlab{}.
\newblock \showarticletitle{Domain Adaptation for Time Series Under Feature and Label Shifts}. In \bibinfo{booktitle}{\emph{International Conference on Machine Learning}}.
\newblock


\bibitem[Hersche et~al\mbox{.}(2022)]%
        {prototype_CL3}
\bibfield{author}{\bibinfo{person}{Michael Hersche}, \bibinfo{person}{Geethan Karunaratne}, \bibinfo{person}{Giovanni Cherubini}, \bibinfo{person}{Luca Benini}, \bibinfo{person}{Abu Sebastian}, {and} \bibinfo{person}{Abbas Rahimi}.} \bibinfo{year}{2022}\natexlab{}.
\newblock \showarticletitle{Constrained few-shot class-incremental learning}. In \bibinfo{booktitle}{\emph{Proceedings of the IEEE/CVF conference on computer vision and pattern recognition}}. \bibinfo{pages}{9057--9067}.
\newblock


\bibitem[Iwasawa and Matsuo(2021)]%
        {T3A}
\bibfield{author}{\bibinfo{person}{Yusuke Iwasawa} {and} \bibinfo{person}{Yutaka Matsuo}.} \bibinfo{year}{2021}\natexlab{}.
\newblock \showarticletitle{Test-time classifier adjustment module for model-agnostic domain generalization}.
\newblock \bibinfo{journal}{\emph{Advances in Neural Information Processing Systems}}  \bibinfo{volume}{34} (\bibinfo{year}{2021}), \bibinfo{pages}{2427--2440}.
\newblock


\bibitem[Jang et~al\mbox{.}(2023)]%
        {TAST}
\bibfield{author}{\bibinfo{person}{Minguk Jang}, \bibinfo{person}{Sae-Young Chung}, {and} \bibinfo{person}{Hye~Won Chung}.} \bibinfo{year}{2023}\natexlab{}.
\newblock \showarticletitle{Test-time adaptation via self-training with nearest neighbor information}. In \bibinfo{booktitle}{\emph{Internetional Conference on Learning Representations}}.
\newblock


\bibitem[Jin et~al\mbox{.}(2022)]%
        {UDA_TS2}
\bibfield{author}{\bibinfo{person}{Xiaoyong Jin}, \bibinfo{person}{Youngsuk Park}, \bibinfo{person}{Danielle Maddix}, \bibinfo{person}{Hao Wang}, {and} \bibinfo{person}{Yuyang Wang}.} \bibinfo{year}{2022}\natexlab{}.
\newblock \showarticletitle{Domain adaptation for time series forecasting via attention sharing}. In \bibinfo{booktitle}{\emph{International Conference on Machine Learning}}. PMLR, \bibinfo{pages}{10280--10297}.
\newblock


\bibitem[Karmanov et~al\mbox{.}(2024)]%
        {TDA}
\bibfield{author}{\bibinfo{person}{Adilbek Karmanov}, \bibinfo{person}{Dayan Guan}, \bibinfo{person}{Shijian Lu}, \bibinfo{person}{Abdulmotaleb El~Saddik}, {and} \bibinfo{person}{Eric Xing}.} \bibinfo{year}{2024}\natexlab{}.
\newblock \showarticletitle{Efficient Test-Time Adaptation of Vision-Language Models}. In \bibinfo{booktitle}{\emph{Proceedings of the IEEE/CVF Conference on Computer Vision and Pattern Recognition}}. \bibinfo{pages}{14162--14171}.
\newblock


\bibitem[Kim et~al\mbox{.}(2023)]%
        {sgem}
\bibfield{author}{\bibinfo{person}{Changhun Kim}, \bibinfo{person}{Joonhyung Park}, \bibinfo{person}{Hajin Shim}, {and} \bibinfo{person}{Eunho Yang}.} \bibinfo{year}{2023}\natexlab{}.
\newblock \showarticletitle{{SGEM}: Test-Time Adaptation for Automatic Speech Recognition via Sequential-Level Generalized Entropy Minimization}. In \bibinfo{booktitle}{\emph{Conference of the International Speech Communication Association (INTERSPEECH)}}.
\newblock


\bibitem[Kim et~al\mbox{.}(2024)]%
        {TTA_AD}
\bibfield{author}{\bibinfo{person}{Dongmin Kim}, \bibinfo{person}{Sunghyun Park}, {and} \bibinfo{person}{Jaegul Choo}.} \bibinfo{year}{2024}\natexlab{}.
\newblock \showarticletitle{When Model Meets New Normals: Test-Time Adaptation for Unsupervised Time-Series Anomaly Detection}. In \bibinfo{booktitle}{\emph{Proceedings of the AAAI Conference on Artificial Intelligence}}, Vol.~\bibinfo{volume}{38}. \bibinfo{pages}{13113--13121}.
\newblock


\bibitem[Kim and Kim(2021)]%
        {tta_speech}
\bibfield{author}{\bibinfo{person}{Sunwoo Kim} {and} \bibinfo{person}{Minje Kim}.} \bibinfo{year}{2021}\natexlab{}.
\newblock \showarticletitle{Test-time adaptation toward personalized speech enhancement: Zero-shot learning with knowledge distillation}. In \bibinfo{booktitle}{\emph{2021 IEEE Workshop on Applications of Signal Processing to Audio and Acoustics (WASPAA)}}. IEEE, \bibinfo{pages}{176--180}.
\newblock


\bibitem[Kim et~al\mbox{.}(2021)]%
        {SFDA}
\bibfield{author}{\bibinfo{person}{Youngeun Kim}, \bibinfo{person}{Donghyeon Cho}, \bibinfo{person}{Kyeongtak Han}, \bibinfo{person}{Priyadarshini Panda}, {and} \bibinfo{person}{Sungeun Hong}.} \bibinfo{year}{2021}\natexlab{}.
\newblock \showarticletitle{Domain adaptation without source data}.
\newblock \bibinfo{journal}{\emph{IEEE Transactions on Artificial Intelligence}} \bibinfo{volume}{2}, \bibinfo{number}{6} (\bibinfo{year}{2021}), \bibinfo{pages}{508--518}.
\newblock


\bibitem[Lee et~al\mbox{.}(2013)]%
        {PL}
\bibfield{author}{\bibinfo{person}{Dong-Hyun Lee} {et~al\mbox{.}}} \bibinfo{year}{2013}\natexlab{}.
\newblock \showarticletitle{Pseudo-label: The simple and efficient semi-supervised learning method for deep neural networks}. In \bibinfo{booktitle}{\emph{Workshop on challenges in representation learning, ICML}}, Vol.~\bibinfo{volume}{3}. Atlanta, \bibinfo{pages}{896}.
\newblock


\bibitem[Lessmeier et~al\mbox{.}(2016)]%
        {mfd_dataset}
\bibfield{author}{\bibinfo{person}{Christian Lessmeier}, \bibinfo{person}{James~Kuria Kimotho}, \bibinfo{person}{Detmar Zimmer}, {and} \bibinfo{person}{Walter Sextro}.} \bibinfo{year}{2016}\natexlab{}.
\newblock \showarticletitle{Condition monitoring of bearing damage in electromechanical drive systems by using motor current signals of electric motors: A benchmark data set for data-driven classification}. In \bibinfo{booktitle}{\emph{PHM Society European Conference}}, Vol.~\bibinfo{volume}{3}.
\newblock


\bibitem[Li et~al\mbox{.}(2017)]%
        {PACS}
\bibfield{author}{\bibinfo{person}{Da Li}, \bibinfo{person}{Yongxin Yang}, \bibinfo{person}{Yi-Zhe Song}, {and} \bibinfo{person}{Timothy~M Hospedales}.} \bibinfo{year}{2017}\natexlab{}.
\newblock \showarticletitle{Deeper, broader and artier domain generalization}. In \bibinfo{booktitle}{\emph{Proceedings of the IEEE international conference on computer vision}}. \bibinfo{pages}{5542--5550}.
\newblock


\bibitem[Li et~al\mbox{.}(2024)]%
        {prototype_CL5}
\bibfield{author}{\bibinfo{person}{Zhuowei Li}, \bibinfo{person}{Long Zhao}, \bibinfo{person}{Zizhao Zhang}, \bibinfo{person}{Han Zhang}, \bibinfo{person}{Di Liu}, \bibinfo{person}{Ting Liu}, {and} \bibinfo{person}{Dimitris~N Metaxas}.} \bibinfo{year}{2024}\natexlab{}.
\newblock \showarticletitle{Steering prototypes with prompt-tuning for rehearsal-free continual learning}. In \bibinfo{booktitle}{\emph{Proceedings of the IEEE/CVF Winter Conference on Applications of Computer Vision}}. \bibinfo{pages}{2523--2533}.
\newblock


\bibitem[Liang et~al\mbox{.}(2024)]%
        {TTA_survey}
\bibfield{author}{\bibinfo{person}{Jian Liang}, \bibinfo{person}{Ran He}, {and} \bibinfo{person}{Tieniu Tan}.} \bibinfo{year}{2024}\natexlab{}.
\newblock \showarticletitle{A comprehensive survey on test-time adaptation under distribution shifts}.
\newblock \bibinfo{journal}{\emph{International Journal of Computer Vision}} (\bibinfo{year}{2024}), \bibinfo{pages}{1--34}.
\newblock


\bibitem[Liang et~al\mbox{.}(2020)]%
        {SHOT_IM}
\bibfield{author}{\bibinfo{person}{Jian Liang}, \bibinfo{person}{Dapeng Hu}, {and} \bibinfo{person}{Jiashi Feng}.} \bibinfo{year}{2020}\natexlab{}.
\newblock \showarticletitle{Do we really need to access the source data? source hypothesis transfer for unsupervised domain adaptation}. In \bibinfo{booktitle}{\emph{International conference on machine learning}}. PMLR, \bibinfo{pages}{6028--6039}.
\newblock


\bibitem[Lin(1991)]%
        {Shannon_entropy}
\bibfield{author}{\bibinfo{person}{Jianhua Lin}.} \bibinfo{year}{1991}\natexlab{}.
\newblock \showarticletitle{Divergence measures based on the Shannon entropy}.
\newblock \bibinfo{journal}{\emph{IEEE Transactions on Information theory}} \bibinfo{volume}{37}, \bibinfo{number}{1} (\bibinfo{year}{1991}), \bibinfo{pages}{145--151}.
\newblock


\bibitem[Lin et~al\mbox{.}(2023)]%
        {video_tta}
\bibfield{author}{\bibinfo{person}{Wei Lin}, \bibinfo{person}{Muhammad~Jehanzeb Mirza}, \bibinfo{person}{Mateusz Kozinski}, \bibinfo{person}{Horst Possegger}, \bibinfo{person}{Hilde Kuehne}, {and} \bibinfo{person}{Horst Bischof}.} \bibinfo{year}{2023}\natexlab{}.
\newblock \showarticletitle{Video test-time adaptation for action recognition}. In \bibinfo{booktitle}{\emph{Proceedings of the IEEE/CVF Conference on Computer Vision and Pattern Recognition}}. \bibinfo{pages}{22952--22961}.
\newblock


\bibitem[Liu et~al\mbox{.}(2023)]%
        {prototype_OS3}
\bibfield{author}{\bibinfo{person}{Jiaming Liu}, \bibinfo{person}{Jun Tian}, \bibinfo{person}{Wei Han}, \bibinfo{person}{Zhili Qin}, \bibinfo{person}{Yulu Fan}, {and} \bibinfo{person}{Junming Shao}.} \bibinfo{year}{2023}\natexlab{}.
\newblock \showarticletitle{Learning multiple gaussian prototypes for open-set recognition}.
\newblock \bibinfo{journal}{\emph{Information Sciences}}  \bibinfo{volume}{626} (\bibinfo{year}{2023}), \bibinfo{pages}{738--753}.
\newblock


\bibitem[Liu and Xue(2021)]%
        {UDA_Align}
\bibfield{author}{\bibinfo{person}{Qiao Liu} {and} \bibinfo{person}{Hui Xue}.} \bibinfo{year}{2021}\natexlab{}.
\newblock \showarticletitle{Adversarial Spectral Kernel Matching for Unsupervised Time Series Domain Adaptation.}. In \bibinfo{booktitle}{\emph{IJCAI}}. \bibinfo{pages}{2744--2750}.
\newblock


\bibitem[Liu et~al\mbox{.}(2021)]%
        {TTT++}
\bibfield{author}{\bibinfo{person}{Yuejiang Liu}, \bibinfo{person}{Parth Kothari}, \bibinfo{person}{Bastien Van~Delft}, \bibinfo{person}{Baptiste Bellot-Gurlet}, \bibinfo{person}{Taylor Mordan}, {and} \bibinfo{person}{Alexandre Alahi}.} \bibinfo{year}{2021}\natexlab{}.
\newblock \showarticletitle{Ttt++: When does self-supervised test-time training fail or thrive?}
\newblock \bibinfo{journal}{\emph{Advances in Neural Information Processing Systems}}  \bibinfo{volume}{34} (\bibinfo{year}{2021}), \bibinfo{pages}{21808--21820}.
\newblock


\bibitem[Lu et~al\mbox{.}(2022)]%
        {prototype_OS1}
\bibfield{author}{\bibinfo{person}{Jing Lu}, \bibinfo{person}{Yunlu Xu}, \bibinfo{person}{Hao Li}, \bibinfo{person}{Zhanzhan Cheng}, {and} \bibinfo{person}{Yi Niu}.} \bibinfo{year}{2022}\natexlab{}.
\newblock \showarticletitle{Pmal: Open set recognition via robust prototype mining}. In \bibinfo{booktitle}{\emph{Proceedings of the AAAI Conference on Artificial Intelligence}}, Vol.~\bibinfo{volume}{36}. \bibinfo{pages}{1872--1880}.
\newblock


\bibitem[Nauta et~al\mbox{.}(2021)]%
        {prototype_Int2}
\bibfield{author}{\bibinfo{person}{Meike Nauta}, \bibinfo{person}{Ron Van~Bree}, {and} \bibinfo{person}{Christin Seifert}.} \bibinfo{year}{2021}\natexlab{}.
\newblock \showarticletitle{Neural prototype trees for interpretable fine-grained image recognition}. In \bibinfo{booktitle}{\emph{Proceedings of the IEEE/CVF conference on computer vision and pattern recognition}}. \bibinfo{pages}{14933--14943}.
\newblock


\bibitem[Niu et~al\mbox{.}(2022)]%
        {EATA}
\bibfield{author}{\bibinfo{person}{Shuaicheng Niu}, \bibinfo{person}{Jiaxiang Wu}, \bibinfo{person}{Yifan Zhang}, \bibinfo{person}{Yaofo Chen}, \bibinfo{person}{Shijian Zheng}, \bibinfo{person}{Peilin Zhao}, {and} \bibinfo{person}{Mingkui Tan}.} \bibinfo{year}{2022}\natexlab{}.
\newblock \showarticletitle{Efficient test-time model adaptation without forgetting}. In \bibinfo{booktitle}{\emph{International conference on machine learning}}. PMLR, \bibinfo{pages}{16888--16905}.
\newblock


\bibitem[Niu et~al\mbox{.}(2023)]%
        {SAR}
\bibfield{author}{\bibinfo{person}{Shuaicheng Niu}, \bibinfo{person}{Jiaxiang Wu}, \bibinfo{person}{Yifan Zhang}, \bibinfo{person}{Zhiquan Wen}, \bibinfo{person}{Yaofo Chen}, \bibinfo{person}{Peilin Zhao}, {and} \bibinfo{person}{Mingkui Tan}.} \bibinfo{year}{2023}\natexlab{}.
\newblock \showarticletitle{Towards Stable Test-Time Adaptation in Dynamic Wild World}. In \bibinfo{booktitle}{\emph{Internetional Conference on Learning Representations}}.
\newblock


\bibitem[Press et~al\mbox{.}(2024)]%
        {Rdumb}
\bibfield{author}{\bibinfo{person}{Ori Press}, \bibinfo{person}{Steffen Schneider}, \bibinfo{person}{Matthias K{\"u}mmerer}, {and} \bibinfo{person}{Matthias Bethge}.} \bibinfo{year}{2024}\natexlab{}.
\newblock \showarticletitle{Rdumb: A simple approach that questions our progress in continual test-time adaptation}.
\newblock \bibinfo{journal}{\emph{Advances in Neural Information Processing Systems}}  \bibinfo{volume}{36} (\bibinfo{year}{2024}).
\newblock


\bibitem[Ragab et~al\mbox{.}(2022)]%
        {UDA_TS3}
\bibfield{author}{\bibinfo{person}{Mohamed Ragab}, \bibinfo{person}{Emadeldeen Eldele}, \bibinfo{person}{Zhenghua Chen}, \bibinfo{person}{Min Wu}, \bibinfo{person}{Chee-Keong Kwoh}, {and} \bibinfo{person}{Xiaoli Li}.} \bibinfo{year}{2022}\natexlab{}.
\newblock \showarticletitle{Self-supervised autoregressive domain adaptation for time series data}.
\newblock \bibinfo{journal}{\emph{IEEE Transactions on Neural Networks and Learning Systems}} (\bibinfo{year}{2022}).
\newblock


\bibitem[Ragab et~al\mbox{.}(2023a)]%
        {adatime}
\bibfield{author}{\bibinfo{person}{Mohamed Ragab}, \bibinfo{person}{Emadeldeen Eldele}, \bibinfo{person}{Wee~Ling Tan}, \bibinfo{person}{Chuan-Sheng Foo}, \bibinfo{person}{Zhenghua Chen}, \bibinfo{person}{Min Wu}, \bibinfo{person}{Chee-Keong Kwoh}, {and} \bibinfo{person}{Xiaoli Li}.} \bibinfo{year}{2023}\natexlab{a}.
\newblock \showarticletitle{Adatime: A benchmarking suite for domain adaptation on time series data}.
\newblock \bibinfo{journal}{\emph{ACM Transactions on Knowledge Discovery from Data}} \bibinfo{volume}{17}, \bibinfo{number}{8} (\bibinfo{year}{2023}), \bibinfo{pages}{1--18}.
\newblock


\bibitem[Ragab et~al\mbox{.}(2023b)]%
        {mapu}
\bibfield{author}{\bibinfo{person}{Mohamed Ragab}, \bibinfo{person}{Emadeldeen Eldele}, \bibinfo{person}{Min Wu}, \bibinfo{person}{Chuan-Sheng Foo}, \bibinfo{person}{Xiaoli Li}, {and} \bibinfo{person}{Zhenghua Chen}.} \bibinfo{year}{2023}\natexlab{b}.
\newblock \showarticletitle{Source-Free Domain Adaptation with Temporal Imputation for Time Series Data}. In \bibinfo{booktitle}{\emph{Proceedings of the 29th ACM SIGKDD Conference on Knowledge Discovery and Data Mining}}. \bibinfo{pages}{1989--1998}.
\newblock


\bibitem[Rebuffi et~al\mbox{.}(2017)]%
        {prototype_CL6}
\bibfield{author}{\bibinfo{person}{Sylvestre-Alvise Rebuffi}, \bibinfo{person}{Alexander Kolesnikov}, \bibinfo{person}{Georg Sperl}, {and} \bibinfo{person}{Christoph~H Lampert}.} \bibinfo{year}{2017}\natexlab{}.
\newblock \showarticletitle{icarl: Incremental classifier and representation learning}. In \bibinfo{booktitle}{\emph{Proceedings of the IEEE conference on Computer Vision and Pattern Recognition}}. \bibinfo{pages}{2001--2010}.
\newblock


\bibitem[Ren and Cheng(2023)]%
        {SFDA_TS2}
\bibfield{author}{\bibinfo{person}{Lei Ren} {and} \bibinfo{person}{Xuejun Cheng}.} \bibinfo{year}{2023}\natexlab{}.
\newblock \showarticletitle{Single/Multi-Source Black-Box Domain Adaption for Sensor Time Series Data}.
\newblock \bibinfo{journal}{\emph{IEEE Transactions on Cybernetics}} (\bibinfo{year}{2023}).
\newblock


\bibitem[Schneider et~al\mbox{.}(2020)]%
        {BN}
\bibfield{author}{\bibinfo{person}{Steffen Schneider}, \bibinfo{person}{Evgenia Rusak}, \bibinfo{person}{Luisa Eck}, \bibinfo{person}{Oliver Bringmann}, \bibinfo{person}{Wieland Brendel}, {and} \bibinfo{person}{Matthias Bethge}.} \bibinfo{year}{2020}\natexlab{}.
\newblock \showarticletitle{Improving robustness against common corruptions by covariate shift adaptation}.
\newblock \bibinfo{journal}{\emph{Advances in neural information processing systems}}  \bibinfo{volume}{33} (\bibinfo{year}{2020}), \bibinfo{pages}{11539--11551}.
\newblock


\bibitem[Snell et~al\mbox{.}(2017)]%
        {prototype_FS1}
\bibfield{author}{\bibinfo{person}{Jake Snell}, \bibinfo{person}{Kevin Swersky}, {and} \bibinfo{person}{Richard Zemel}.} \bibinfo{year}{2017}\natexlab{}.
\newblock \showarticletitle{Prototypical networks for few-shot learning}.
\newblock \bibinfo{journal}{\emph{Advances in neural information processing systems}}  \bibinfo{volume}{30} (\bibinfo{year}{2017}).
\newblock


\bibitem[Su et~al\mbox{.}(2022)]%
        {TTAC}
\bibfield{author}{\bibinfo{person}{Yongyi Su}, \bibinfo{person}{Xun Xu}, {and} \bibinfo{person}{Kui Jia}.} \bibinfo{year}{2022}\natexlab{}.
\newblock \showarticletitle{Revisiting realistic test-time training: Sequential inference and adaptation by anchored clustering}.
\newblock \bibinfo{journal}{\emph{Advances in Neural Information Processing Systems}}  \bibinfo{volume}{35} (\bibinfo{year}{2022}), \bibinfo{pages}{17543--17555}.
\newblock


\bibitem[Su et~al\mbox{.}(2024)]%
        {TRIBE}
\bibfield{author}{\bibinfo{person}{Yongyi Su}, \bibinfo{person}{Xun Xu}, {and} \bibinfo{person}{Kui Jia}.} \bibinfo{year}{2024}\natexlab{}.
\newblock \showarticletitle{Towards Real-World Test-Time Adaptation: Tri-Net Self-Training with Balanced Normalization}.
\newblock \bibinfo{journal}{\emph{The 38th AAAI Conference on Artificial Intelligence}} (\bibinfo{year}{2024}).
\newblock


\bibitem[Sukhbaatar et~al\mbox{.}(2015)]%
        {cross_entropy_reason2}
\bibfield{author}{\bibinfo{person}{Sainbayar Sukhbaatar}, \bibinfo{person}{Joan Bruna}, \bibinfo{person}{Manohar Paluri}, \bibinfo{person}{Lubomir Bourdev}, {and} \bibinfo{person}{Rob Fergus}.} \bibinfo{year}{2015}\natexlab{}.
\newblock \showarticletitle{Training convolutional networks with noisy labels}.
\newblock \bibinfo{journal}{\emph{International Conference on Learning Representations}} (\bibinfo{year}{2015}).
\newblock


\bibitem[Sun et~al\mbox{.}(2020)]%
        {TTT}
\bibfield{author}{\bibinfo{person}{Yu Sun}, \bibinfo{person}{Xiaolong Wang}, \bibinfo{person}{Zhuang Liu}, \bibinfo{person}{John Miller}, \bibinfo{person}{Alexei Efros}, {and} \bibinfo{person}{Moritz Hardt}.} \bibinfo{year}{2020}\natexlab{}.
\newblock \showarticletitle{Test-time training with self-supervision for generalization under distribution shifts}. In \bibinfo{booktitle}{\emph{International conference on machine learning}}. PMLR, \bibinfo{pages}{9229--9248}.
\newblock


\bibitem[Tsai et~al\mbox{.}(2024)]%
        {GDA}
\bibfield{author}{\bibinfo{person}{Yun-Yun Tsai}, \bibinfo{person}{Fu-Chen Chen}, \bibinfo{person}{Albert~YC Chen}, \bibinfo{person}{Junfeng Yang}, \bibinfo{person}{Che-Chun Su}, \bibinfo{person}{Min Sun}, {and} \bibinfo{person}{Cheng-Hao Kuo}.} \bibinfo{year}{2024}\natexlab{}.
\newblock \showarticletitle{GDA: Generalized Diffusion for Robust Test-time Adaptation}. In \bibinfo{booktitle}{\emph{Proceedings of the IEEE/CVF Conference on Computer Vision and Pattern Recognition}}. \bibinfo{pages}{23242--23251}.
\newblock


\bibitem[Um et~al\mbox{.}(2017)]%
        {magnitude_warp_aug}
\bibfield{author}{\bibinfo{person}{Terry~T Um}, \bibinfo{person}{Franz~MJ Pfister}, \bibinfo{person}{Daniel Pichler}, \bibinfo{person}{Satoshi Endo}, \bibinfo{person}{Muriel Lang}, \bibinfo{person}{Sandra Hirche}, \bibinfo{person}{Urban Fietzek}, {and} \bibinfo{person}{Dana Kuli{\'c}}.} \bibinfo{year}{2017}\natexlab{}.
\newblock \showarticletitle{Data augmentation of wearable sensor data for parkinson’s disease monitoring using convolutional neural networks}. In \bibinfo{booktitle}{\emph{Proceedings of the 19th ACM international conference on multimodal interaction}}. \bibinfo{pages}{216--220}.
\newblock


\bibitem[Wang et~al\mbox{.}(2020)]%
        {TENT}
\bibfield{author}{\bibinfo{person}{Dequan Wang}, \bibinfo{person}{Evan Shelhamer}, \bibinfo{person}{Shaoteng Liu}, \bibinfo{person}{Bruno Olshausen}, {and} \bibinfo{person}{Trevor Darrell}.} \bibinfo{year}{2020}\natexlab{}.
\newblock \showarticletitle{Tent: Fully test-time adaptation by entropy minimization}.
\newblock \bibinfo{journal}{\emph{arXiv preprint arXiv:2006.10726}} (\bibinfo{year}{2020}).
\newblock


\bibitem[Wang et~al\mbox{.}(2022)]%
        {CoTTA}
\bibfield{author}{\bibinfo{person}{Qin Wang}, \bibinfo{person}{Olga Fink}, \bibinfo{person}{Luc Van~Gool}, {and} \bibinfo{person}{Dengxin Dai}.} \bibinfo{year}{2022}\natexlab{}.
\newblock \showarticletitle{Continual test-time domain adaptation}. In \bibinfo{booktitle}{\emph{Proceedings of the IEEE/CVF Conference on Computer Vision and Pattern Recognition}}. \bibinfo{pages}{7201--7211}.
\newblock


\bibitem[Wang et~al\mbox{.}(2023)]%
        {TSD}
\bibfield{author}{\bibinfo{person}{Shuai Wang}, \bibinfo{person}{Daoan Zhang}, \bibinfo{person}{Zipei Yan}, \bibinfo{person}{Jianguo Zhang}, {and} \bibinfo{person}{Rui Li}.} \bibinfo{year}{2023}\natexlab{}.
\newblock \showarticletitle{Feature alignment and uniformity for test time adaptation}. In \bibinfo{booktitle}{\emph{Proceedings of the IEEE/CVF Conference on Computer Vision and Pattern Recognition}}. \bibinfo{pages}{20050--20060}.
\newblock


\bibitem[Wei et~al\mbox{.}(2023)]%
        {prototype_CL4}
\bibfield{author}{\bibinfo{person}{Yujie Wei}, \bibinfo{person}{Jiaxin Ye}, \bibinfo{person}{Zhizhong Huang}, \bibinfo{person}{Junping Zhang}, {and} \bibinfo{person}{Hongming Shan}.} \bibinfo{year}{2023}\natexlab{}.
\newblock \showarticletitle{Online prototype learning for online continual learning}. In \bibinfo{booktitle}{\emph{Proceedings of the IEEE/CVF International Conference on Computer Vision}}. \bibinfo{pages}{18764--18774}.
\newblock


\bibitem[Wilson and Cook(2020)]%
        {UDA}
\bibfield{author}{\bibinfo{person}{Garrett Wilson} {and} \bibinfo{person}{Diane~J Cook}.} \bibinfo{year}{2020}\natexlab{}.
\newblock \showarticletitle{A survey of unsupervised deep domain adaptation}.
\newblock \bibinfo{journal}{\emph{ACM Transactions on Intelligent Systems and Technology (TIST)}} \bibinfo{volume}{11}, \bibinfo{number}{5} (\bibinfo{year}{2020}), \bibinfo{pages}{1--46}.
\newblock


\bibitem[Wilson et~al\mbox{.}(2020)]%
        {UDA_TS4}
\bibfield{author}{\bibinfo{person}{Garrett Wilson}, \bibinfo{person}{Janardhan~Rao Doppa}, {and} \bibinfo{person}{Diane~J Cook}.} \bibinfo{year}{2020}\natexlab{}.
\newblock \showarticletitle{Multi-source deep domain adaptation with weak supervision for time-series sensor data}. In \bibinfo{booktitle}{\emph{Proceedings of the 26th ACM SIGKDD international conference on knowledge discovery \& data mining}}. \bibinfo{pages}{1768--1778}.
\newblock


\bibitem[Wilson et~al\mbox{.}(2023)]%
        {UDA_TS5}
\bibfield{author}{\bibinfo{person}{Garrett Wilson}, \bibinfo{person}{Janardhan~Rao Doppa}, {and} \bibinfo{person}{Diane~J Cook}.} \bibinfo{year}{2023}\natexlab{}.
\newblock \showarticletitle{Calda: Improving multi-source time series domain adaptation with contrastive adversarial learning}.
\newblock \bibinfo{journal}{\emph{IEEE Transactions on Pattern Analysis and Machine Intelligence}} (\bibinfo{year}{2023}).
\newblock


\bibitem[Xia et~al\mbox{.}(2023)]%
        {prototype_OS4}
\bibfield{author}{\bibinfo{person}{Ziheng Xia}, \bibinfo{person}{Penghui Wang}, \bibinfo{person}{Ganggang Dong}, {and} \bibinfo{person}{Hongwei Liu}.} \bibinfo{year}{2023}\natexlab{}.
\newblock \showarticletitle{Adversarial kinetic prototype framework for open set recognition}.
\newblock \bibinfo{journal}{\emph{IEEE Transactions on Neural Networks and Learning Systems}} (\bibinfo{year}{2023}).
\newblock


\bibitem[Xiao et~al\mbox{.}(2023)]%
        {SFDA_TS}
\bibfield{author}{\bibinfo{person}{Yutang Xiao}, \bibinfo{person}{Hongbo Shi}, \bibinfo{person}{Bing Song}, \bibinfo{person}{Yang Tao}, \bibinfo{person}{Shuai Tan}, {and} \bibinfo{person}{Boyu Wang}.} \bibinfo{year}{2023}\natexlab{}.
\newblock \showarticletitle{Temporal Attention Source-Free Adaptation for Chemical Processes Fault Diagnosis}.
\newblock \bibinfo{journal}{\emph{IEEE Transactions on Industrial Informatics}} (\bibinfo{year}{2023}).
\newblock


\bibitem[Xu et~al\mbox{.}(2020)]%
        {prototype_OS5}
\bibfield{author}{\bibinfo{person}{Wenjia Xu}, \bibinfo{person}{Yongqin Xian}, \bibinfo{person}{Jiuniu Wang}, \bibinfo{person}{Bernt Schiele}, {and} \bibinfo{person}{Zeynep Akata}.} \bibinfo{year}{2020}\natexlab{}.
\newblock \showarticletitle{Attribute prototype network for zero-shot learning}.
\newblock \bibinfo{journal}{\emph{Advances in Neural Information Processing Systems}}  \bibinfo{volume}{33} (\bibinfo{year}{2020}), \bibinfo{pages}{21969--21980}.
\newblock


\bibitem[Yang et~al\mbox{.}(2020)]%
        {prototype_OS2}
\bibfield{author}{\bibinfo{person}{Hong-Ming Yang}, \bibinfo{person}{Xu-Yao Zhang}, \bibinfo{person}{Fei Yin}, \bibinfo{person}{Qing Yang}, {and} \bibinfo{person}{Cheng-Lin Liu}.} \bibinfo{year}{2020}\natexlab{}.
\newblock \showarticletitle{Convolutional prototype network for open set recognition}.
\newblock \bibinfo{journal}{\emph{IEEE Transactions on Pattern Analysis and Machine Intelligence}} \bibinfo{volume}{44}, \bibinfo{number}{5} (\bibinfo{year}{2020}), \bibinfo{pages}{2358--2370}.
\newblock


\bibitem[Yuan et~al\mbox{.}(2023)]%
        {RoTTA}
\bibfield{author}{\bibinfo{person}{Longhui Yuan}, \bibinfo{person}{Binhui Xie}, {and} \bibinfo{person}{Shuang Li}.} \bibinfo{year}{2023}\natexlab{}.
\newblock \showarticletitle{Robust test-time adaptation in dynamic scenarios}. In \bibinfo{booktitle}{\emph{Proceedings of the IEEE/CVF Conference on Computer Vision and Pattern Recognition}}. \bibinfo{pages}{15922--15932}.
\newblock


\bibitem[Yue et~al\mbox{.}(2021)]%
        {prototype_FS4}
\bibfield{author}{\bibinfo{person}{Xiangyu Yue}, \bibinfo{person}{Zangwei Zheng}, \bibinfo{person}{Shanghang Zhang}, \bibinfo{person}{Yang Gao}, \bibinfo{person}{Trevor Darrell}, \bibinfo{person}{Kurt Keutzer}, {and} \bibinfo{person}{Alberto~Sangiovanni Vincentelli}.} \bibinfo{year}{2021}\natexlab{}.
\newblock \showarticletitle{Prototypical cross-domain self-supervised learning for few-shot unsupervised domain adaptation}. In \bibinfo{booktitle}{\emph{Proceedings of the IEEE/CVF Conference on Computer Vision and Pattern Recognition}}. \bibinfo{pages}{13834--13844}.
\newblock


\bibitem[Zhang et~al\mbox{.}(2021)]%
        {prototype_FS2}
\bibfield{author}{\bibinfo{person}{Baoquan Zhang}, \bibinfo{person}{Xutao Li}, \bibinfo{person}{Yunming Ye}, \bibinfo{person}{Zhichao Huang}, {and} \bibinfo{person}{Lisai Zhang}.} \bibinfo{year}{2021}\natexlab{}.
\newblock \showarticletitle{Prototype completion with primitive knowledge for few-shot learning}. In \bibinfo{booktitle}{\emph{Proceedings of the IEEE/CVF conference on computer vision and pattern recognition}}. \bibinfo{pages}{3754--3762}.
\newblock


\bibitem[Zhang et~al\mbox{.}(2022)]%
        {MEMO}
\bibfield{author}{\bibinfo{person}{Marvin Zhang}, \bibinfo{person}{Sergey Levine}, {and} \bibinfo{person}{Chelsea Finn}.} \bibinfo{year}{2022}\natexlab{}.
\newblock \showarticletitle{Memo: Test time robustness via adaptation and augmentation}.
\newblock \bibinfo{journal}{\emph{Advances in Neural Information Processing Systems}}  \bibinfo{volume}{35} (\bibinfo{year}{2022}), \bibinfo{pages}{38629--38642}.
\newblock


\bibitem[Zhang et~al\mbox{.}(2023)]%
        {AaDNPC}
\bibfield{author}{\bibinfo{person}{Yifan Zhang}, \bibinfo{person}{Xue Wang}, \bibinfo{person}{Kexin Jin}, \bibinfo{person}{Kun Yuan}, \bibinfo{person}{Zhang Zhang}, \bibinfo{person}{Liang Wang}, \bibinfo{person}{Rong Jin}, {and} \bibinfo{person}{Tieniu Tan}.} \bibinfo{year}{2023}\natexlab{}.
\newblock \showarticletitle{Adanpc: Exploring non-parametric classifier for test-time adaptation}. In \bibinfo{booktitle}{\emph{International Conference on Machine Learning}}. PMLR, \bibinfo{pages}{41647--41676}.
\newblock


\bibitem[Zhang and Sabuncu(2018)]%
        {cross_entropy_reason1}
\bibfield{author}{\bibinfo{person}{Zhilu Zhang} {and} \bibinfo{person}{Mert Sabuncu}.} \bibinfo{year}{2018}\natexlab{}.
\newblock \showarticletitle{Generalized cross entropy loss for training deep neural networks with noisy labels}.
\newblock \bibinfo{journal}{\emph{Advances in neural information processing systems}}  \bibinfo{volume}{31} (\bibinfo{year}{2018}).
\newblock


\bibitem[Zhou et~al\mbox{.}(2023)]%
        {prototype_FS3}
\bibfield{author}{\bibinfo{person}{Fei Zhou}, \bibinfo{person}{Peng Wang}, \bibinfo{person}{Lei Zhang}, \bibinfo{person}{Wei Wei}, {and} \bibinfo{person}{Yanning Zhang}.} \bibinfo{year}{2023}\natexlab{}.
\newblock \showarticletitle{Revisiting prototypical network for cross domain few-shot learning}. In \bibinfo{booktitle}{\emph{Proceedings of the IEEE/CVF Conference on Computer Vision and Pattern Recognition}}. \bibinfo{pages}{20061--20070}.
\newblock


\bibitem[Zhu et~al\mbox{.}(2021)]%
        {prototype_CL1}
\bibfield{author}{\bibinfo{person}{Fei Zhu}, \bibinfo{person}{Xu-Yao Zhang}, \bibinfo{person}{Chuang Wang}, \bibinfo{person}{Fei Yin}, {and} \bibinfo{person}{Cheng-Lin Liu}.} \bibinfo{year}{2021}\natexlab{}.
\newblock \showarticletitle{Prototype augmentation and self-supervision for incremental learning}. In \bibinfo{booktitle}{\emph{Proceedings of the IEEE/CVF Conference on Computer Vision and Pattern Recognition}}. \bibinfo{pages}{5871--5880}.
\newblock


\end{thebibliography}

\appendix
\section{Appendix}
\subsection{Algorithm of ACCUP}
\label{algorithm_1}
\begin{algorithm}
\caption{Augmented Contrastive Clustering with Uncertainty-Aware
Prototyping for Time Series Test Time Adaptation}\label{algorithm}
\SetKwInOut{Input}{Input}\SetKwInOut{Output}{Output}
\SetKw{Return}{Return}
\Input{Feature encoder $f_\theta$, classifier head $h_\phi$, test batch $\mathbb{B}$}
\Output{Predictions for all $x \in \mathbb{B}$}
Initialize the support set $\mathbb{S}$ for each category using the weights parameter of the source domain classifier\;
\For{streamed batch data $x$ in $\mathbb{B}$}{
 \tcp{obtain augmentation-ensemble logits}
 $p_{ens} = compute\_aug\_ensemble\_logits(x)$ (Eq.1)\\
 \tcp{update support set and obtain uncertainty-aware prototypes}
 $\mu = compute\_class\_prototypes(\mathbb{S})$ (Eq.2)\\
 \tcp{obtain prototype-based logits}
 $p_{proto} = compute\_prototype\_logits(\mu, p_{ens})$ (Eq.3) \\
 \tcp{obtain reliable pseudo labels}
 $p_{out} = entropy\_comparison(p_{ens}, p_{proto})$ (Eq.4)\\
 \tcp{optimize by augmented contrastive clustering}
 $\mathcal{L} = \sum_{i=1}^{n_B} \mathcal{L}_{contrast}^i$ (Eq.5)\\
 update network $f_{\theta}$ to minimize $\mathcal{L}$
 }
 \Return{
 $p_{out}$ for all $x \in \mathbb{B}$
 }
\end{algorithm}

\subsection{The Effects of Using Different Weightings for Augmentation-Ensemble}
\label{appendix_different_weight_ensemble}
To investigate the impact of different weightings between raw and augmented data on model performance, we conducted experiments on three datasets using both fixed and learnable weight combinations. For fixed weights, we varied the raw data weight from $0.1$ to $0.9$ in $0.1$ increments, with the augmented data weight correspondingly adjusted. For learnable weights, the model learned optimal weights during training. Results presented in Table \ref{tab_different_weight_ensemble} indicate that while learnable weights did not yield significant improvements due to the limited parameter update steps in the TTA setting, manually tuned weights demonstrated performance variations across datasets. For instance, a raw data weight of $0.2$ optimized performance on UCIHAR, while $0.1$ was optimal for SSC. These findings suggest that emphasizing magnitude warping through increased augmentation weights can potentially enhance model performance. However, overall, the use of different weight combinations had a limited impact on the final model performance.
\begin{table}[h]
\centering
\caption{Performance analysis of using different weighting for augmentation-ensembles. The sum of the weights of the raw and augmented samples is one.}
\begin{NiceTabular}{@{}c|ccc@{}} 
\toprule 
Weights on $f_{raw}$ and $p_{raw}$ & UCIHAR & MFD & SSC \\ \midrule
0.1 & 88.35 & 95.58 & \textbf{62.71}  \\
0.2 & \textbf{88.37} & \textbf{95.60} & 62.57  \\
0.3 & 88.11 & \textbf{95.60} & 62.63  \\
0.4 & 88.09 & \textbf{95.60} & 62.63  \\
0.5 (avg) & 88.16 & \textbf{95.60} & 62.65 \\
0.6 & 88.16 & \textbf{95.60} & 62.67 \\
0.7 & 87.56 & 95.59 & 62.55 \\
0.8 & 87.48 & 95.59 & 62.55 \\
0.9 & 87.44 & 95.58 & 62.42 \\
\midrule
learnable\tabularnote{The learnable weight is initialized with $0.5$} & 88.11 & \textbf{95.60} & 62.48 \\
\bottomrule
\end{NiceTabular}
\label{tab_different_weight_ensemble}
\end{table}

\subsection{Implementation Details}
\label{implementation_details}
\subsubsection{Feature Encoder Backbone} This study adopts the same feature extraction backbone as prior works \cite{adatime, CA-TCC, mapu}. The architecture is a one-dimensional convolutional neural network consisting of three convolutional layers. The filter sizes of each convolutional layer are 64, 128, and 128, respectively. Rectified linear unit nonlinear activation and batch normalization modules are used behind each convolutional layer.
\subsubsection{Model Parameters} Our model incorporates three main parameters: $K$, the number of samples per class retained in the memory support set after entropy-based filtering, $\eta$, the scaling factor for uncertainty-aware prototypes,  and $\tau$, the temperature parameter in augmented contrastive clustering. In our configuration, the parameter settings vary across different datasets. For the $K$ parameter, we use a setting of 10 for the UCIHAR dataset, 100 for the MFD dataset, and 50 for the SSC dataset. The scaling parameter $\eta$, is set at 20 for UCIHAR, 1 for MFD, and 50 for SSC. Lastly, the temperature parameter $\tau$, is determined to be 0.7 for UCIHAR, 0.6 for MFD, and 0.3 for SSC, catering to the specific requirements of each dataset.
\subsubsection{Training Scheme} For a fair comparison with baselines, all methods share the same feature encoder backbone and training procedure. Hyperparameter tuning for baselines involved evaluating 20 random combinations from a defined search space to identify optimal configurations. We ensured a level playing field by using the same pre-trained model on the source domain for all compared methods. The source-domain model was pre-trained for 40 epochs on all datasets using a batch size of 32 and a learning rate of 1e-3 with Adam optimization. During test-time adaptation, the learning rate was set to 1e-5 for SSC and 3e-4 for UCIHAR and MFD. Each new batch of samples triggers a prediction and then a single gradient descent step for the model update. We built all the experiments using Pytorch and trained them on an NVIDIA GeForce RTX 2080Ti GPU. The macro F1 score was chosen as the evaluation metric for its robustness with imbalanced data. To account for randomness, results represent the mean and standard deviation of three consecutive runs per cross-domain scenario.

\subsection{Descriptions of Datasets}
\label{intro_datasets}
\subsubsection{UCIHAR Dataset}
The UCIHAR dataset \cite{har_dataset} is designed specifically for tasks related to human activity recognition. Data collection involved the utilization of three distinct sensor types: accelerometer sensor, gyroscope sensor, and body sensor. Each sensor yielded three-dimensional readings, resulting in a total of 9 channels per sample, with 128 data points per sample. The dataset encompasses information gathered from 30 distinct users, with each user representing an individual domain. In our experimental design, we executed five cross-user experiments, wherein the model was trained on one user and subsequently tested on different users, facilitating an assessment of its cross-domain performance.
\subsubsection{Machine Fault Diagnosis (MFD) Dataset}
The MFD dataset ~\cite{mfd_dataset}, curated by Paderborn University, is intended for fault diagnosis applications, employing vibration signals to discern various types of incipient faults. The data collection encompasses four distinct working conditions, with each data sample comprising a solitary univariate channel containing 5120 data points, as established in previous studies ~ \cite{UDA_TS3, mapu}. In our experimental framework, each working condition is treated as an individual domain, and we implement five diverse cross-condition scenarios to assess the domain adaptation performance.
\subsubsection{Sleep Stage Classification (SSC) Dataset}
The SSC task involves the classification of EEG signals into five distinct stages: Wake (W), Non-Rapid Eye Movement stages (N1, N2, N3), and Rapid Eye Movement (REM). This study employs the Sleep-EDF dataset \cite{ssc_dataset}, which encompasses EEG recordings obtained from 20 healthy subjects. Following the methodology of prior research \cite{mapu}, we focus on a single channel, specifically Fpz-Cz, and leverage data from 10 subjects to design five cross-domain experiments.
\subsubsection{PACS Dataset}
\label{intro_pacs_dataset}
The PACS dataset \cite{PACS} comprises $9991$ RGB images of size $224x224$, categorized into seven classes across four domains: Photo (P), Art painting (A), Cartoon (C), and Sketch (S). Following the established protocol in previous research \cite{T3A, TSD}, we configure four cross-domain scenarios to evaluate domain adaptation performance, with each domain serving as the target domain in turn while the remaining domains constitute the source domains.

Table \ref{tbl:datasets} provides a detailed overview of three real-world time series datasets, including the number of domains, sensor channels, classes, sample length, and total samples in the training and testing splits.
\begin{table}[!tbh]
\centering
\caption{Details of the adopted datasets (C: \#channels, K: \#classes, L: sample length).}
\begin{NiceTabular}{l|ccc|cc}
\toprule
\textbf{Dataset} & \textbf{C}  & \textbf{K} & \textbf{L} & \# train samples & \# test samples \\ \midrule

UCIHAR &  9 & 6 & 128 & 2300 & 990 \\ 
MFD & 1 & 3 & 5120 & 7312 & 3604 \\ 
SSC & 1 & 5 & 3000 & 14280 & 6130 \\ 
\bottomrule
\end{NiceTabular}
\label{tbl:datasets}
\end{table}

\begin{table*}[t]
\centering
\caption{Comparison results with advanced TTA methods with ResNet18 backbone on four PACS cross-domain scenarios.}
\setlength{\tabcolsep}{2.5mm}{
\begin{NiceTabular}{@{}l|cccc|c@{}} 
\toprule 
Algorithm & A & C & P & S &  Average\\ \midrule
Source & 80.56$\pm$0.45 & 77.36$\pm$0.85 & 93.01$\pm$0.17 & 77.35$\pm$2.90 & 82.07$\pm$0.49 \\
BN \cite{BN} & 81.60$\pm$0.16 & 82.00$\pm$0.51 & 92.85$\pm$0.24 & 74.86$\pm$1.10 & 82.82$\pm$0.34 \\
TENT \cite{TENT} & 83.43$\pm$0.53 & 83.02$\pm$0.74 & 93.88$\pm$0.38 & 79.35$\pm$1.15 & 84.92$\pm$0.32 \\
PL \cite{PL} & 84.93$\pm$1.32 & 83.27$\pm$1.78 & 92.62$\pm$1.37 & 77.72$\pm$3.66 & 84.64$\pm$1.13  \\
SHOT-IM \cite{SHOT_IM} & 84.61$\pm$1.06 & 82.36$\pm$1.88 & 93.60$\pm$0.38 & 69.64$\pm$3.40 & 82.55$\pm$1.07  \\
T3A \cite{T3A} & 83.00$\pm$0.76 & 79.56$\pm$0.44 & 94.48$\pm$0.34 & 76.95$\pm$2.94 & 83.50$\pm$0.67 \\
EATA \cite{EATA} & 81.33$\pm$0.30 & 81.89$\pm$0.55 & 92.82$\pm$0.53 & 74.77$\pm$0.91 & 82.70$\pm$0.31 \\
LAME \cite{LAME} & 83.05$\pm$0.53 & 83.06$\pm$0.51 & 94.30$\pm$0.29 & 77.91$\pm$0.80 & 84.58$\pm$0.23 \\
TSD \cite{TSD} & \textbf{86.50$\pm$0.75} & \underline{86.38$\pm$0.82} & \underline{94.57$\pm$0.32} & \underline{81.84$\pm$0.94} & \underline{87.32$\pm$0.39} \\
\midrule
\textbf{Proposed}  & \underline{86.23$\pm$1.35} & \textbf{87.12$\pm$1.63} & \textbf{95.87$\pm$0.51} & \textbf{84.12$\pm$0.54} & \textbf{88.33$\pm$0.31} \\ 
\bottomrule
\end{NiceTabular}}
\label{tab_pacs_resnet18}
\end{table*}



\begin{table*}[t]
\centering
\caption{Comparison results with advanced TTA methods with ResNet50 backbone on four PACS cross-domain scenarios.}
\setlength{\tabcolsep}{2.5mm}{
\begin{NiceTabular}{@{}l|cccc|c@{}} 
\toprule 
Algorithm & A & C & P & S &  Average\\ \midrule
Source & 82.50$\pm$1.83 & 80.80$\pm$0.33 & 94.05$\pm$0.30 & 80.99$\pm$1.29 & 84.59$\pm$0.40 \\
BN \cite{BN} & 83.27$\pm$0.47 & 84.91$\pm$0.43 & 94.03$\pm$0.31 & 77.92$\pm$1.23 & 85.03$\pm$0.20 \\
TENT \cite{TENT} & 85.28$\pm$1.07 & 86.75$\pm$0.92 & 94.94$\pm$0.83 & 82.96$\pm$1.20 & 87.48$\pm$0.52 \\
PL \cite{PL} & 83.96$\pm$1.63 & 84.15$\pm$2.91 & 93.82$\pm$1.74 & 78.99$\pm$2.64 & 85.23$\pm$1.70  \\
SHOT-IM \cite{SHOT_IM} & 84.31$\pm$0.63 & 85.74$\pm$0.56 & 94.04$\pm$0.67 & 77.91$\pm$0.94 & 85.50$\pm$0.31  \\
T3A \cite{T3A} & 84.07$\pm$0.68 & 82.37$\pm$0.92 & 95.02$\pm$0.27 & 82.72$\pm$1.06 & 86.04$\pm$0.24 \\
EATA \cite{EATA} & 83.27$\pm$0.47 & 84.91$\pm$0.43 & 94.03$\pm$0.31 & 77.92$\pm$1.24 & 85.04$\pm$0.20 \\
LAME \cite{LAME} & 84.97$\pm$0.77 & 85.50$\pm$0.55 & 95.04$\pm$0.23 & 80.97$\pm$1.09 & 86.62$\pm$0.22 \\
TSD \cite{TSD} & \underline{87.68$\pm$0.84} & \textbf{88.78$\pm$0.63} & \underline{96.17$\pm$0.37} & \underline{85.01$\pm$1.52} & \underline{89.41$\pm$0.51} \\
\midrule
\textbf{Proposed}  & \textbf{89.42$\pm$0.72} & \underline{87.46$\pm$1.23} & \textbf{96.99$\pm$0.33} & \textbf{86.23$\pm$0.25} & \textbf{90.03$\pm$0.43} \\ 
\bottomrule
\end{NiceTabular}}
\label{tab_pacs_resnet50}
\end{table*}



\subsection{Descriptions of Baseline Methods} \label{intro_baseline_methods}
\begin{itemize}
    \item \textbf{Target} indicates that the upper bound of domain adaptation, where training and testing are performed on the target domain only.
    \item \textbf{Source} indicates that no adaptation is performed and only the source domain model is used to inference on the target domain. 
    \item \textbf{BN} \cite{BN} Test time normalization updates the statistics of the batch normalization module by using streamed data from the target domain. 
    \item \textbf{PL} \cite{PL} Pseudo Label facilitates the parameter updating of all normalization layers by minimizing the cross-entropy loss associated with the predicted pseudo labels.
    \item \textbf{TENT} ~\cite{TENT} Test time entropy minimization facilitates the parameter updating of all normalization layers by minimizing the entropy of the model predictions in the streamed data. 
    \item \textbf{SHOT} \cite{SHOT_IM} Source hypothesis transfer involves learning the target-specific feature extraction module by leveraging both information maximization and self-supervised pseudo-labeling techniques. 
    \item \textbf{EATA} ~\cite{EATA} Efficient anti-forgetting test-time adaptation employs a sample-efficient entropy minimization loss to exclude uninformative samples and introduces a regularization loss designed to preserve critical model weights throughout the adaptation process.
    \item \textbf{SAR} \cite{SAR} introduces a sharpness-aware and reliable entropy minimization method to stabilize the TTA process.
    \item \textbf{TTAC} \cite{TTAC} Test-Time Anchored Clustering mitigates the KL divergence between the distributions of the source and target domains. 
    \item \textbf{NOTE} \cite{NOTE} Non-i.i.d. Test-Time Adaptation (NOTE) optionally updates batch normalization statistics when the distributional shift between a test instance and the source model's training distribution exceeds a predefined threshold.
    \item \textbf{T3A} \cite{T3A} Test time template adjuster creates a pseudo-prototype for each class using streamed target domain data and then classifies each sample based on its distance to the pseudo-prototype. 
    \item \textbf{TAST} \cite{TAST} Test-time adaptation via self-training with nearest neighbor information introduces trainable adaptation modules atop the feature extractor, computing estimated class distributions through the aggregation of pseudo labels derived from nearest neighbor information. 
    \item \textbf{CoTTA} \cite{CoTTA} Continual Test-Time Adaptation implements a mean-teacher architecture and randomly selects and restores the model parameters to those of the source model. 
    \item \textbf{RoTTA} \cite{RoTTA} Robust Test-Time Adaptation incorporates the replacement of batch normalization layers with Robust Batch Normalization, enhancing the accuracy of estimating target domain batch normalization statistics. 
    \item \textbf{TSD} \cite{TSD} Test Time self-distillation employs self-distillation to enhance uniformity in the target feature during adaptation. Additionally, it incorporates memorized spatial local clustering to incentivize the alignment of feature representations in the latent space with the pseudo logits.
\end{itemize}

\subsection{Generalization on Vision Task}
\label{appendix_generalization_vision_tasks}
To assess the model's generalizability, we conducted additional experiments on the PACS dataset\cite{PACS}, a benchmark for visual domain adaptation. The more description of PACS dataset is presented in Appendix \ref{intro_pacs_dataset}. The model was adapted to the image domain by replacing the time series-specific feature encoder with ResNet18 and ResNet50 architectures, respectively. Data augmentation techniques commonly used in image processing (random cropping, flipping, and color jittering) were employed as counterparts to the time series-specific magnitude warping. To ensure fair comparison, all experimental settings were consistent with previous studies, and we directly quoted baseline results from prior research \cite{TSD}. For source training, we used an Adam optimizer with a learning rate of 5e-5 and a batch size of 32. For test-time adaptation, we employed an Adam optimizer with a learning rate of 5e-5 and a batch size of 128. All experiments were conducted using three random seeds $\{0, 1, 2\}$, and we report the average accuracies in Tables \ref{tab_pacs_resnet18} and \ref{tab_pacs_resnet50}.

The results demonstrate that our proposed model consistently outperforms baseline approaches. Specifically, with the ResNet18 backbone, we achieved optimal performance in three of the four cross-domain scenarios, attaining the highest average accuracy of 88.33\% – approximately a 1\% improvement over the second-best baseline. Using the ResNet50 backbone, we achieved optimal performance in three of the four cross-domain scenarios, with an overall classification performance exceeding 90\%. These findings underscore the model's versatility potential and the effectiveness of our proposed uncertainty-aware prototypical ensemble approach in enhancing model adaptability beyond time series domains.

\subsection{Ablation Study of Each Module}
\label{appendix_ablation_of_each_module}
To assess the individual contributions of each model component, ablation study was conducted by removing specific modules. Table \ref{tab_ablation_each_module} presents classification performance with the following ablation configurations: a model without contrast learning (w/o Contrast), without entropy comparison (w/o entComp), without augmentation (w/o augmentation), and without prototype learning (w/o prototypes). Results indicate that removing different modules decreases model performance to varying degrees, with the relative contribution of modules differing slightly across datasets. For instance, the uncertainty-aware prototype significantly improves performance on UCIHAR and SSC by 1.34\% and 4.29\%, respectively. On the MFD dataset, augmentated contrastive clustering and entropy comparison strategies has a greater impact than other modules, improving performance by 2.56\% and 1.18\%, respectively. Notably, the \textit{w/o prototype} model variant achieved optimal results on MFD. This may be attributed to the fact that for data with longer sequence lengths and fewer categories, classification-based prediction logits exhibit better discriminative properties due to richer information content. These findings underscore the importance of each component in achieving optimal model performance while highlighting the need for dataset-specific considerations in module selection and optimization.

\begin{table}[h]
\centering
\caption{Ablation experiments of removing each module.}
\setlength{\tabcolsep}{4.5mm}{
\begin{NiceTabular}{@{}l|ccc@{}} 
\toprule 
Augmentations & UCIHAR & MFD & SSC \\ \midrule
w/o contrast & 87.34 & 93.04 & 62.54  \\ 
w/o entComp & 87.24 & 94.42  & 58.67  \\ 
w/o augmentation & 87.79 & 95.10 & 62.47 \\
w/o prototypes & 86.82 & \textbf{95.64} & 58.36   \\
\midrule
\textbf{Proposed} & \textbf{88.16} & 95.60 & \textbf{62.65} \\
\bottomrule
\end{NiceTabular}}
\label{tab_ablation_each_module}
\end{table}


\subsection{Effectiveness of Different Augmentation Combination Strategies}
\label{appendix_different_augmentation_combination}
We conduct additional experiments to explore augmentation combinations in Table \ref{tab_augmentation_combination}. The results suggest that while combining augmentations can enhance performance, our method—whether used alone or in conjunction with baseline methods—consistently outperforms others. For instance, using our method with Jitter augmentation achieved an 88.48\% F1-score on the UCIHAR dataset, whereas our method alone scored 95.60\% on the MFD dataset. These results further indicate that the time series-specific amplitude and noise variations can be effectively mitigated through the use of a suitable augmentation strategy, enhancing the robustness of the model during the test adaptation phase.

\begin{table}[h]
    \centering
    \caption{Experiment results with different augmentation combinations on three datasets}
    \begin{tabular}{cccc}
    \toprule
        Augmentations & UCIHAR & MFD & SSC  \\ \midrule
        Jitter+Permutation & 81.32 & 94.93 & 63.03 \\
        Scale+Permutation & 81.48 & 94.59 & 63.08 \\
        Jitter+Scale & 88.29 & 95.53 & 62.80 \\
        Jitter+Scale+Permutation & 79.71 & 94.65 & \textbf{63.20} \\
        \midrule
        Permutation+Ours & 80.62 & 94.71 & 63.14 \\
        Scale+Ours & 88.17 & 95.43 & 62.77 \\
        Jitter+Ours & \textbf{88.48} & 94.75 & 62.65 \\
        {Ours} & 88.16 & \textbf{95.60} & 62.65 \\
        \bottomrule
    \end{tabular}
\label{tab_augmentation_combination} 
\end{table}

\subsection{Analysis of Fine-Tuing Different Layers of Feature Encoder} \label{fine_tune_different_layers}
This study examines the impact of fine-tuning feature encoder parameters across various layers on model performance. Table ~ \ref{tab_finetune_different_layers} presents the mean and standard deviation of macro F1 scores achieved when fine-tuning different layers during adaptation. Fine-tuning all feature encoding layers yielded the best classification results, suggesting a significant improvement in the model's discriminative power through modifications across all layers. Interestingly, classification performance demonstrated resilience, showing only a slight decline when adjusting parameters for select layers, compared to fine-tuning all layers. This observation implies that competitive classification results can still be achieved by fine-tuning specific layers only, highlighting the model's flexibility and robustness to parameter adjustments at different levels of the feature encoding hierarchy.

\begin{table}[h]
\centering
\caption{Performance analysis of fine-tuning the parameters of different layers of the model.}
\setlength{\tabcolsep}{4.5mm}{
\begin{NiceTabular}{@{}c|ccc@{}} 
\toprule 
Layers & UCIHAR & MFD & SSC \\ \midrule
Conv1 & 87.93 & 95.08 & 62.27 \\
Conv2 & 87.77 & 95.33 & 62.29 \\
Conv3 & 87.91 & 95.26 & \underline{62.49} \\
Conv1 \& 2 & 87.99 & 95.30 & 60.99 \\
Conv1 \& 3 & \underline{88.04} & \underline{95.44} & 62.17 \\
Conv2 \& 3 & 87.79 & 95.39 & 61.97 \\ \midrule
Conv1 \& 2 \& 3 & \textbf{88.16} & \textbf{95.60}  & \textbf{62.65} \\
\bottomrule
\end{NiceTabular}}
\label{tab_finetune_different_layers}
\end{table}


\end{document}